\documentclass[preprint]{elsarticle}
\usepackage{graphicx}
\usepackage{tabularx}
\usepackage[linesnumbered,ruled,vlined]{algorithm2e}
\usepackage{amssymb}
\usepackage{subfigure}
\usepackage{array}
\usepackage{lscape}
\usepackage{dcolumn,longtable,hhline,colortbl}
\usepackage[rotateright]{rotating}
\usepackage{booktabs,makecell,multirow}
\usepackage[section]{placeins}
\usepackage{amsmath}
\usepackage{enumerate}
\usepackage{color,colortbl}
\usepackage{hyperref}
\usepackage{lineno}
\usepackage[margin=2cm]{geometry}
\usepackage{color}
\usepackage{soul}
\usepackage{url}
\usepackage[table]{xcolor}
\usepackage{makecell}
\usepackage{amsthm}
\usepackage{multirow}
\usepackage{bm}
\usepackage{autoaligne}
\usepackage{amsmath}
\usepackage{mathtools}
\usepackage{booktabs}

\usepackage[flushleft]{threeparttable}
\usepackage{setspace}

\usepackage{nomencl}
\makenomenclature

\usepackage[fontsize=10.5pt]{scrextend}
\usepackage{multirow}

\SetKwComment{Comment}{/* }{ */}

\renewcommand\theadalign{bc}
\renewcommand\theadfont{\bfseries}
\renewcommand\theadgape{\Gape[4pt]}
\renewcommand\cellgape{\Gape[4pt]}

\makeatletter
\hypersetup{
    colorlinks=true,
    linkcolor=blue,
    filecolor=magenta,
    urlcolor=cyan,
}

\journal{Reliability Engineering \& System Safety}

\begin{document}

\begin{frontmatter}

\title{A generic physics-informed neural network-based framework for reliability assessment of multi-state systems\footnote{The source codes of this paper will be made publicly accessible at GitHub: \url{https://github.com/zxgcqupt/PINNs4SRA} upon the publication of this paper.}}

\author[address1]{Taotao Zhou}
\author[address2]{Xiaoge Zhang\corref{label1}}
\ead{xiaoge.zhang@polyu.edu.hk}
\author[address3,address4]{Enrique Lopez Droguett}
\author[address4]{Ali Mosleh}

\cortext[label1]{Corresponding author.}

\address[address1]{Center for Risk and Reliability, University of Maryland, College Park, MD, USA}
\address[address2]{Department of Industrial and Systems Engineering, The Hong Kong Polytechnic University, Kowloon, Hong Kong}
\address[address3]{Department of Civil and Environmental Engineering, University of California, Los Angeles, CA, USA}
\address[address4]{Garrick Institute for the Risk Sciences, University of California, Los Angeles, CA, USA}

\begin{abstract}
In this paper, we leverage the recent advances in physics-informed neural network (PINN) and develop a generic PINN-based framework to assess the reliability of multi-state systems (MSSs). The proposed methodology consists of two major steps. In the first step, we recast the reliability assessment of MSS as a machine learning problem using the framework of PINN. A feedforward neural network with two individual loss groups are constructed to encode the initial condition and state transitions governed by ordinary differential equations (ODEs) in MSS. Next, we tackle the problem of high imbalance in the magnitude of the back-propagated gradients in PINN from a multi-task learning perspective. Particularly, we treat each element in the loss function as an individual task, and adopt a gradient surgery approach named projecting conflicting gradients (PCGrad), where a task's gradient is projected onto the norm plane of any other task that has a conflicting gradient. The gradient projection operation significantly mitigates the detrimental effects caused by the gradient interference when training PINN, thus accelerating the convergence speed of PINN to high-precision solutions to MSS reliability assessment. With the proposed PINN-based framework, we investigate its applications for MSS reliability assessment in several different contexts in terms of time-independent or dependent  state transitions and system scales varying from small to medium. The results demonstrate that the proposed PINN-based framework shows generic and remarkable performance in MSS reliability assessment, and the incorporation of PCGrad in PINN leads to substantial improvement in solution quality and convergence speed.
\end{abstract}

\end{frontmatter}

\section{Introduction}
Reliability assessment and optimization of engineered systems have received growing attention in a broad range of sectors, such as power grid~\cite{dehghani2021adaptive,al2012novel}, transportation systems~\cite{edrissi2015transportation,zhang2017game,zhang2021bayesian}, computing systems~\cite{mo2015performability,mo2017performability}, electrical and mechanical systems~\cite{xu2019reliability,zafar2020efficient,moustafa2021system,vohra2020fast}. Over the last few years, the increasing occurrences of extreme events have posed more than ever pressing needs for highly reliable infrastructure systems so that they can still operate at a desirable performance under extreme natural conditions. The malfunction and failure of these critical systems can lead to catastrophic consequences in terms of economic loss and human fatality. Take the 2021 Texas power crisis as an example, the inadequately winterizing of power equipment significantly compromised the reliability of power transmission system and resulted in the partial failure of the power grid, which eventually led to a massive power outage and left 4.5 million homes and businesses without power for several days~\cite{2021_texas_crisis}.

In the context of reliability assessment and optimization of engineered systems, one of the popular means in characterizing engineered systems is to model it as a multi-state system (MSS)~\cite{rausand2003system,lisnianski2003multi,yingkui2012multi,xie2004computing}. Differing from the binary-state reliability models, which assumes that a system and its components only have two states (i.e., perfectly operational or complete failure), MSS introduces a finite number of intermediate states for each system component to indicate a wide range of performance levels that lie between the perfectly functioning state and the completely failed state~\cite{lisnianski2003multi}. The rich intermediate states in MSS models enable the representation of the deterioration behavior of engineered systems in a finer granularity than that of traditional binary-state system models. As a consequence, MSS models have become an appealing tool for modelling and assessing system reliability in a broad array of industrial applications. For example,~\citet{qiu2019reliability} modeled a power distribution system as a MSS and developed a universal generating function (UGF)-based approach to quantify its reliability, where power transmission loss was taken into consideration.~\citet{liu2019reliability} modeled the stochastic dependency among state transitions of a MSS or component via copula functions and studied the reliability assessment of MSS with state transition dependency.~\citet{iscioglu2021reliability} evaluated the reliability of a MSS consisting of $n$ identical independent units with two different types of dependency among components.~\citet{mi2018reliability} developed an evidential network-based method to analyze the reliability of complex MSS with epistemic uncertainty. 


In general, the methods used for reliability assessment of MSS can be roughly categorized into five classes: stochastic process particularly Markov process method~\cite{lisnianski2012multi,liu2006reliability}, extensions of conventional binary reliability model such as multi-state fault tree method~\cite{janan1985multistate}, Monte Carlo simulation (MCS) method~\cite{ramirez2005composite,zeng2021resilience}, universal generating function (UGF) method~\cite{levitin2005universal}, and Bayesian network~\cite{si2010integrated}. Among them, MCS is one of the most popularly used approaches for system reliability assessment owing to its easiness to implement, advantages in uncertainty representation and propagation as well as flexibility in characterizing complex system behavior and interactions among system components. For example,~\citet{zio2004estimation} exploited the flexibility of MCS and developed quantitative measures to estimate the importance of components in a multi-state series-parallel system.~\citet{echard2011ak} developed an efficient active learning method that combined Kriging with Monte Carlo Simulation to perform reliability assessment in structural systems.~\citet{schneider2013social} treated social network as a multi-state commodity and applied reliability measures commonly used in MSS to quantify the influence of a given actor in the social network. 

In principle, the reliability of MSS can be accurately estimated using the standard MCS method as long as sufficient MCS samples are generated. However, despite the popularity of MCS, its computational effort grows exponentially in accordance with the number of components and component-wise states in MSS. The number of MCS samples needed to estimate the reliability of MSS in large-scale systems at a high precision easily gets computationally unaffordable. Such a flaw in MCS makes it inapplicable in time-sensitive applications that require real-time decision-making support. One alternative approach is to build data-driven surrogate models for the concerned MSS by taking advantage of the recent advances in deep learning. Unfortunately, deep learning faces similar issues when dealing with MSS in the high-dimensional space. To be specific, a considerable amount of data needs to be collected to represent MSS in a wide range of scenarios (e.g., different degradation conditions, deterioration trajectories) in order to train a deep learning model. The collection of such a representative training dataset for MSS can take a long time and might incur unaffordable costs in some cases. 

A promising direction to address the aforementioned issues is to encode physical laws (or empirical laws) in the development of machine learning models, which is referred to as Physics-Informed Machine Learning (PIML) in the literature~\cite{raissi2019physics,karniadakis2021physics,lu2021deepxde}. A representative example along this front is the family of Physics-Informed Neural Networks (PINNs)~\cite{raissi2019physics}. The physical laws (i.e., conservation laws) governing system behaviors in the form of partial differential equation (PDE) or ordinary differential equation (ODE) are usually rigorously derived from first principles. In PINN, physical laws are typically incorporated as a soft loss term in the objective function of deep learning models. The incorporation of physical laws in the deep learning models substantially prunes the parameter search space as solutions violating the physical laws are discarded immediately. As a result, encoding physical laws in machine learning models essentially reduces the number of training points that are required to tune a deep learning model. The benefits of exploiting physical laws in building efficient deep learning models have been showcased in several recent studies~\cite{kapusuzoglu2021information,zhao2021physics,chao2022fusing,zobeiry2021physics,zhou2021physics,cofre2021remaining}.  It is worthwhile noting that the loss functions of PINNs is complicated and involve multiple terms, which would compete with each other during the training process~\cite{karniadakis2021physics}. Hence, since PINNs is a highly non-convex optimization problem, it is essential to assure the stability and robustness in the training of PINNs, which remains an active research topic yet~\cite{wang2022and}.

The application of PINN in MSS reliability assessment has been rarely studied in the literature even though several features of MSS make it a natural fit to be formulated and solved as a PINN-type problem. Specifically, the stochastic behavior of component state transitions in MSS is commonly characterized as Markov process~\cite{dui2015semi,lisnianski2017recent,barbu2017semi,eryilmaz2015dynamic}, which are usually difficult to derive analytical solutions. Numerical methods are typically adopted, such as differential equation solver and Monte Carlo simulation. These numerical methods are prohibitively computational expensive, and they get computationally unaffordable easily when extensive uncertainty and sensitivity analysis are needed. The ODE governing the Markov processes~\cite{freidlin1996markov} makes PINN a viable solution for MSS reliability assessment. Secondly, existing approaches often discretize the life span of MSS into multiple equally-sized time intervals. The side effect of doing this is that the reliability of MSS can only be performed in the pre-specified discrete time instants. In contrast, the development of PINN for MSS reliability assessment frees us from the MSS life span discretization, and it allows to estimate the reliability of MSS at any time instant in a mesh-free fashion.

To address the above issues, in this paper, we are motivated to develop a generic framework casting reliability assessment of MSS as a machine learning problem by exploiting the power of PINN. Towards this goal, one common pain point in adopting PINN is that the original formulation by~\citet{raissi2019physics} often struggles to approximate the exact solution of PDEs in high precision due to the extremely imbalanced gradients during the training of PINN via back-propagation~\cite{karniadakis2021physics,wang2022and,wang2021understanding}. To address the imbalanced gradients among loss terms in PINN, we treat each loss term as an individual task and tackle this problem from a multi-task learning perspective following the approach proposed by~\citet{yu2020gradient}. The key idea is to project a task’s gradient onto the normal plane of the gradient of any other task that has a conflicting gradient. Compared to previous studies, we make the following contributions:
\begin{enumerate}
    \item Formulation of a generic physics-informed neural network-based framework to tackle the system reliability assessment problem in MSS. The developed PINN-based framework provides a novel and effective paradigm for assessing the reliability of complex MSS. 
    
    \item To address the issue of the extremely imbalanced loss function in PINNs, we integrate the gradient surgery method with PINN to deconflict gradients during the training of PINN via back-propagation. The incorporation of the gradient surgery approach in PINN significantly accelerates the convergence speed of PINN and substantially improves the solution quality in MSS reliability assessment. 
    
    \item We investigate the applications of the PINN-based framework in several different scenarios in terms of system state transition rates (e.g., homogeneous continuous-time Markov chain and non-homogeneous continuous-time Markov chain) and system scales (e.g., small-scale MSS, medium-scale MSS). Besides, we examine the quality of the solutions from PINN by comparing them with that of Matlab solver.   
\end{enumerate}

The rest of the paper is structured as follows. Section 2 provides a brief introduction to the multi-state systems (MSS) and describes the technical background for PINN. Section 3 develops the proposed methodology to build PINN for MSS reliability assessment. Section 4 shows the applications of PINN in MSS reliability assessment and compares its performance with two other alternatives. Section 5 ends this paper with concluding remarks and discusses future research directions.

\section{Background}
In this section, we briefly introduce the technical background of reliability modeling of multi-state system and the mathematical formulation of physics-informed neural network. 

\subsection{MSS Reliability Model}\label{sec:MSS_reliability}
Traditional binary reliability models only allow two operational states: perfectly functioning or complete failure. Whereas, MSS reliability assessment associates the system and its components with multiple intermediate states as indicated by either performance capacity or damage severity during performance degradation. Suppose the performance of a MSS is characterized by $M + 1$ discrete ordered states~\cite{rausand2003system}, represented by the following set:
\begin{equation}
{S} = \left\{ {0,1, \cdots ,M} \right\}  
\end{equation}
where $0$ denotes the worst state, and $M$ denotes the best state. The others are intermediate states between the worst and the best states. 

Suppose the probabilities associated with the $M + 1$ states in the MSS at time $t$ is denoted by the following vector:
\begin{equation}
\bm{p}\left( t \right) = \left[ {{p_0}\left( t \right),{p_1}\left( t \right), \cdots ,{p_M}\left( t \right)} \right]
\end{equation}

As the probability vector $\bm{p}\left( t \right)$ constitutes the exhaustive set of all the MSS states, it needs to satisfy the following constraint:
\begin{equation}
\sum\limits_{i = 0}^M {{p_i}\left( t \right)}  = 1,\;\forall t \in \left[ {1,2, \cdots ,T} \right]
\end{equation}
where $T$ denotes the MSS operation period.

\begin{figure}[!ht]
    \centering
    \includegraphics[scale=0.7]{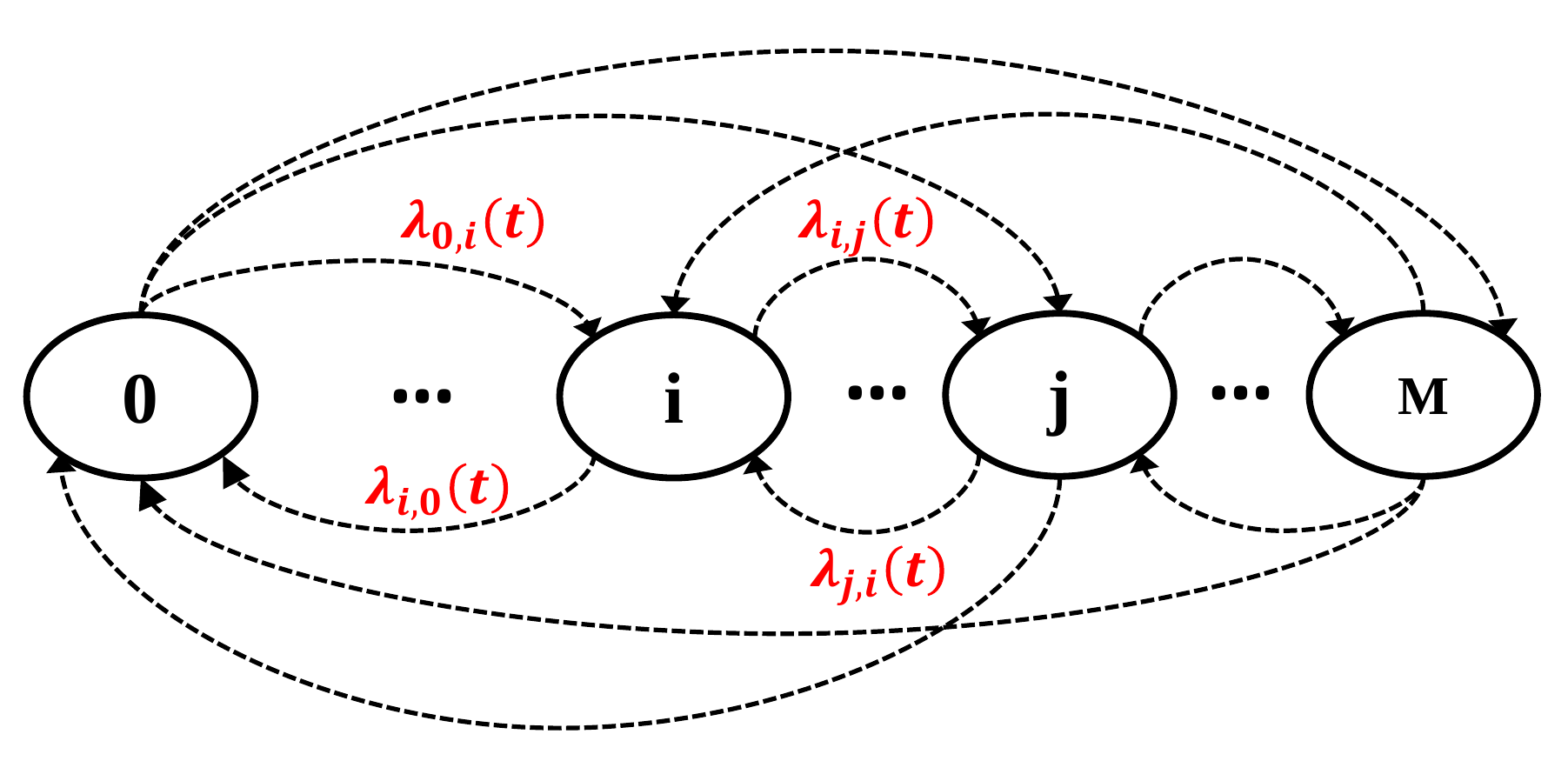}
    \caption{Graphical representation of state transitions in a multi-state system}
    \label{fig:state_transition}
\end{figure}

In general, the system dynamics in a MSS at each time instant $t$ is characterized by a state-transition diagram as shown in Fig.~\ref{fig:state_transition}. Each node in Fig.~\ref{fig:state_transition} represents the probability associated with the state $i$, and each branch labeled ${\lambda _{i,j}}\left( t \right)$ denotes the corresponding one-step transition probability from state $i$ to state $j$ at the time instant $t$. Mathematically, the state transition probabilities among all the states at the time instant $t$ can be represented by the following matrix:
\begin{equation}\label{eq:transition}
\bm{Q}\left( t \right) = \left[ {\begin{array}{*{20}{c}}
{{\lambda _{0,0}}\left( t \right)}&{{\lambda _{0,1}}\left( t \right)}& \cdots &{{\lambda _{0,M}}\left( t \right)}\\
{{\lambda _{1,0}}\left( t \right)}&{{\lambda _{1,1}}\left( t \right)}& \cdots &{{\lambda _{1,M}}\left( t \right)}\\
 \vdots & \vdots & \ddots & \vdots \\
{{\lambda _{M,0}}\left( t \right)}&{{\lambda _{M,1}}\left( t \right)}& \cdots &{{\lambda _{M,M}}\left( t \right)}
\end{array}} \right]
\end{equation}
where ${\lambda _{i,i}}\left( t \right) =  - \sum\limits_{j = 0,j \ne i}^M {{\lambda _{i,j}}\left( t \right),} \quad \forall i \in S$. Obviously, the sum of all the elements in each row is zero. 

With the properly defined state transition matrix $\bm{Q}\left( t \right)$, then the evolution of the states in MSS over time can be described using the Kolmogorov forward equation:
\begin{equation}\label{eq:transition_equation}
\begin{array}{l}
{\bm{p}^{'}}\left( t \right) = \bm{p}\left( t \right)\bm{Q}\left( t \right)\\
\bm{p}\left( {t = 0} \right) = {\bm{s}_0}
\end{array}
\end{equation}
where ${\bm{p}^{'}}\left( t \right)$ refers to the first-order derivative of $\bm{p}\left( t \right)$ at the time instant $t$ and $\bm{p}\left( {t = 0} \right)$ denotes the initial system state at the time instant $0$. 

Given the initial system state $\bm{s}_0$, MSS evolves over time following the state transition matrix $\bm{Q}\left( t \right)$. The MSS reliability can be derived by aggregating the state probability associated with system states that perform its desired function during the mission time. Mathematically, it is formulated as below:
\begin{equation}\label{eq:system_reliability}
R = \sum\limits_{i = 0}^M {{\delta _i}{p_i}\left( t \right)} 
\end{equation}
where $\delta_i$ is a binary variable indicating whether state $i$ satisfies the desired property at the system level. If ${\delta _i} = 1$, then state $i$ meets the intended function, otherwise, state $i$ does not meet the requirement; ${{p_i}\left( t \right)}$ denotes the probability of state $i$ at the time instant $t$.

\subsection{Physics-Informed Neural Networks}
In this section, we explain the underlying architecture of physics-informed neural networks (PINNs) and describes the mathematical formulation of PINNs. 

In several industrial applications, the behavior of dynamical systems are described by general nonlinear partial differential equation (PDE)~\cite{karniadakis2021physics,pang2019fpinns}. Consider a PDE represented in a general form formulated in Eq. (\ref{eq:pde}):
\begin{equation}\label{eq:pde}
{\bm{u}\left( {\bm{x},t} \right)} + \mathcal{N}_{\bm{x}}\left[ \bm{u} \right] = 0,\;\;\bm{x} \in \Omega ,\;\;t \in \left[ {0,T} \right]
\end{equation}
where $\bm{u}\left( {\bm{x},t} \right)$ denotes the latent solution, $\mathcal{N}_{\bm{x}}\left[ \bullet \right]$ is a nonlinear differential operator, $\bm{x}$ denotes a vector of space coordinates, and $t$ denotes the time. The domain $\Omega$ of the PDE is bounded based on the prior knowledge of the dynamic systems, and $\left[ {0,T} \right]$ is the time interval within which the system evolves. 

It is well-known that neural networks are universal function approximators to learn the unknown relationship between any inputs and outputs. As a result, neural networks can be used to approximate the solution to the PDE function shown in Eq. (\ref{eq:pde}). Suppose we denote the left-hand-side of Eq. (\ref{eq:pde}) as $f\left( {t,\bm{x}} \right)$:
\begin{equation}\label{eq:reformulation}
f:{\bm{u}\left( {\bm{x},t} \right)} + \mathcal{\bm{N}}_{\bm{x}}\left[ \bm{u} \right]
\end{equation}

$f\left( {t,\bm{x}} \right)$ now acts as a constraint modelling the physical law described by the PDE in Eq. (\ref{eq:pde}). The first term $u\left( {\bm{x},t} \right)$ in Eq. (\ref{eq:pde}) can be approximated by a neural network, where $\bm{x}$ and $t$ are inputs to the neural network. The neural network for approximating $u\left( {\bm{x},t} \right)$ together with Eq. (\ref{eq:reformulation}) (here, Eq. (\ref{eq:reformulation}) acts as an equality constraint) result in a physics-informed neural network. Regarding the nonliner differentiator $\mathcal{\bm{N}}_{\bm{x}}\left[ \bm{u} \right]$, its value can be derived using the same neural network that is used to approximate $u\left( {\bm{x},t} \right)$, where automatic differentiation can be applied to differentiate compositions of functions following the chain rule~\cite{baydin2018automatic}. 

The neural network approximating $u\left( {\bm{x},t} \right)$ has the same parameters as the network representing $f\left( {t,\bm{x}} \right)$. The weights of the neural network can be optimized by minimizing the following function:
\begin{equation}\label{eq:total_loss}
{\mathcal{L}_{MSE}} = \mathcal{L}_{MSE}^u + \mathcal{W}  * \mathcal{L}_{MSE}^f
\end{equation}
where $\mathcal{W}$ is a factor denoting the weight associated with the loss term $\mathcal{L}_{MSE}^f$, and
\begin{equation}
\mathcal{L}_{MSE}^u = \frac{1}{{{N_u}}}{\sum\limits_{i = 1}^{{N_u}} {\left[ {u_{NN}\left( {t_u^i,x_u^i; \bm{\theta}} \right) - {u^i}} \right]} ^2}
\end{equation}
and
\begin{equation}
\mathcal{L}_{MSE}^f = \frac{1}{{{N_f}}}{\sum\limits_{i = 1}^{{N_f}} {\left[ {f\left( {t_f^i,x_f^i} \right)} \right]} ^2}
\end{equation}
where $\left\{ {t_u^i,x_u^i,{u^i}} \right\}$ denotes the initial and boundary data points on $u\left( {t,x} \right)$, ${u_{NN}\left( {t_u^i,x_u^i; \bm{\theta}} \right)}$ denotes the prediction of the neural network on the inputs ${\left( {t_u^i,x_u^i} \right)}$, $\bm{\theta}$ refers to the weights in the neural network, $\left\{ {t_f^i,x_f^i} \right\}$ represents the collocation points for $f\left( {t,\bm{x}} \right)$, $N_u$ and $N_f$ represent the number of points generated for $u\left( {\bm{x}, t} \right)$ and $f\left( {t,\bm{x}} \right)$, respectively. 

In Eq. (\ref{eq:total_loss}), $\mathcal{L}_{MSE}^u$ measures the loss of the neural network when approximating the function $u\left( {\bm{x}, t} \right)$, while $\mathcal{L}_{MSE}^f$ enforces the physical law imposed by Eq. (\ref{eq:pde}) into the neural network via a series of collocation points ${\left( {t_f^i,x_f^i} \right)}$. With the training data consisting of boundary and collocation points, we seek to minimize the loss function formulated in Eq. (\ref{eq:total_loss}) through optimizing the weights in the neural network via gradient descent algorithms. As illustrated in Eq. (\ref{eq:total_loss}), PINNs provide a rigorous way to seamlessly integrate the information from both the measurement data and physical laws, where physical laws are encoded into the loss function of the neural network via automatic differentiation. Consideration of underlying physical laws prunes the feasible solution space to the neural network parameters, and thus significantly reduces the number of training points as well as the size of the neural network (e.g., number of layers, number of hidden node in each layer etc.) during model training. 

\section{Proposed Framework}\label{sec:proposed_methodology}
In this section, we introduce the proposed PINN-based framework for MSS reliability assessment in details. The proposed framework consists of two major steps. In the first step, we recast MSS reliability assessment as a machine learning problem in the framework of PINN. Next, we outline the gradient surgery approach to minimize gradient conflicts among multiple tasks during the training of PINN.

\subsection{PINNs for MSS Reliability Assessment}
As introduced in Section~\ref{sec:MSS_reliability}, there are two key components in the reliability assessment of a MSS that need to be appropriately characterized in the framework of PINN. The first key component is the initial state $\bm{s}_0$ denoting the states of the MSS at the time instant $0$. The second core component is the state transition, which is described by the Kolmogorov forward equations as shown in Eq. (\ref{eq:transition_equation}).

\begin{figure}[!ht]
    \centering
    \includegraphics[scale=0.55]{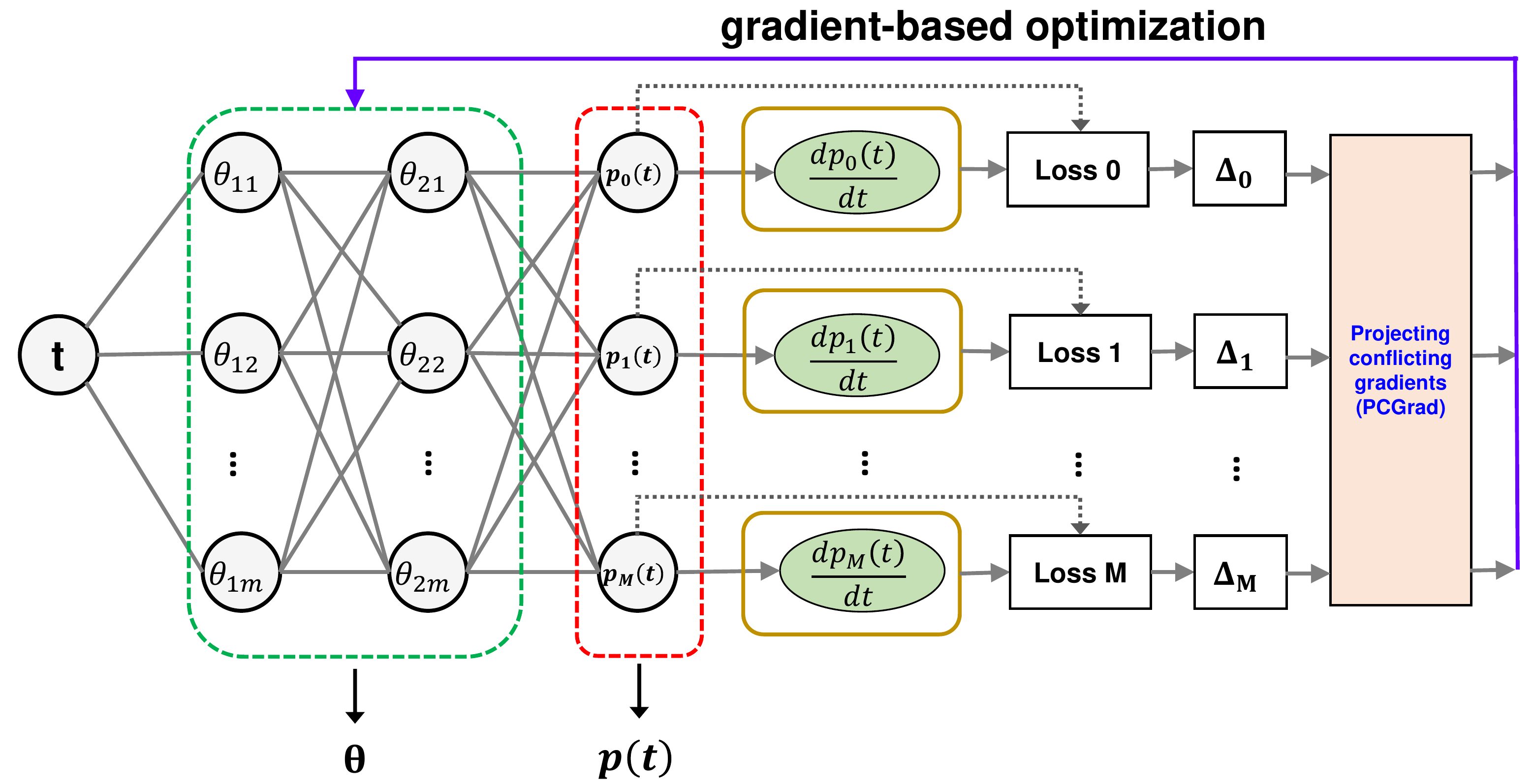}
    \caption{Configuration of the proposed PINN-based framework for multi-state system reliability assessment}
    \label{fig:architecture}
\end{figure}

For the sake of demonstration, Fig. \ref{fig:architecture} illustrates the configuration of a PINN composed of two hidden layers with each layer having $m$ hidden units for reliability assessment in MSS. In practice, a PINN can consist of as many hidden layers and hidden units as needed. Suppose we discretize the operation horizon $\left[ {1,2, \cdots ,T} \right]$ of a MSS into $T$ time steps, which is typically referred to as collocation points.  Each time we feed a specific time step $t$ into the PINN, we obtain the probability associated with each state in $\bm{p}\left( t \right)$. With automatic differentiation, we derive the first order derivative of $\bm{p}\left( t \right)$ corresponding to each state with respect to the time instant $t$. Next, we compute the loss function between the first order derivative of $\bm{p}\left( t \right)$ and the exact solution $\bm{p}\left( t \right)\bm{Q}\left( t \right)$ as illustrated in Eq. (\ref{eq:transition_equation}). 
 
In addition to incorporating the differential equations characterizing state transitions, another constraint that needs to be modeled is the initial state. Basically, at the time instant $0$, the system starts at a specific condition characterized by the probability associated with each state in MSS. Suppose $s_0^i$ denotes the probability corresponding to state $i$ at the time instant $0$, then, in conjunction with the loss function for state transition modeling, we have the loss function for the MSS reliability assessment as below: 
\begin{equation}\label{eq:mss_loss}
\mathcal{L}\left( \bm{\theta}  \right) = \underbrace{\sum\limits_{i = 0}^M {{{\left[ {u_{NN}^i\left( {t = 0;\bm{\theta} } \right) - s_0^i} \right]}^2}}}_{\mathclap{\mathcal{L}_u: \text{Loss on initial condition}}}  + \mathcal{W} \underbrace{\left[ {\frac{1}{T}\sum\limits_{t = 1}^T { {{{\left( {u_{NN}\left( {t;\bm{\theta} } \right)\bm{Q}\left( t \right) - \frac{{du_{NN}\left( {t;\bm{\theta} } \right)}}{{dt}}} \right)}^2}} } } \right]}_{\mathclap{\mathcal{L}_f: \text{Loss on approximating state transition equation}}}
\end{equation}
where $M$ refers to the number of states in the MSS, $T$ is the number of time steps, ${u_{NN}}\left( \bm{\theta}  \right)$ denotes the neural network with its weights represented by the parameter $\bm{\theta}$, ${u_{NN}^i\left( {t = 0;\bm{\theta} } \right)}$ indicates the probability corresponding to state $i$ at the time instant $0$ as predicted by the neural network, $\mathcal{W}$ is a weighting factor, and ${\frac{{du_{NN}^i\left( {t;\bm{\theta} } \right)}}{{dt}}}$ reveals the derivative with respect to the time $t$ that is estimated by the neural network ${u_{NN}}\left( \bm{\theta}  \right)$ via automatic differentiation. 

As shown in Eq. (\ref{eq:mss_loss}), there are two key components in the loss function $\mathcal{L}\left( \bm{\theta}  \right)$. The first component $\mathcal{L}_u$ uses a mean squared error metric to evaluate the loss corresponding to the initial states of the MSS, while the second component $\mathcal{L}_f$ enforces the structure modeling of the state transition in MSS and
estimates the residual when approximating the governing equation characterizing the state transitions in MSS. Combining the two loss terms together, the goal is to minimize the loss function $\mathcal{L}\left( \bm{\theta}  \right)$ through optimizing the parameter $\bm{\theta}$. With a set of training data, we can reduce the loss of the neural network iteratively via backpropgation using gradient descent algorithms, such as Adam~\cite{jais2019adam}.

After the PINN is properly trained, then it can be used to estimate the probability corresponding to any state at any given time instant $t$. Once the probability associated with each state is accurately predicted, the reliability of MSS at system level can be inferred following Eq. (\ref{eq:system_reliability}). Different from existing methods, a significant advantage of PINNs is that they allow to estimate the reliability of MSS in a continuous manner. Most of existing approaches can only estimate the probability associated with each state at predetermined time instants, while PINN is mesh-free and it allows to tackle MSS reliability assessment in a comprehensive fashion. 

\subsection{Conflicting Gradients Projection for Physics-Informed Neural Networks}
As reported by~\citet{wang2021understanding}, PINN faces a fundamental mode of failure that is closely related to the stiffness of the back-propagated gradient flows because the loss terms $\mathcal{L}_u$ and $\mathcal{L}_f$ are highly imbalanced in magnitude. In particular, the loss term $\mathcal{L}_f$ characterizing the PDE residual dominates the loss function and, consequently, the optimization algorithm is heavily biased towards minimizing the loss term $\mathcal{L}_f$. As a result, PINN performs poorly in fitting the initial conditions, and leads to quite unstable and erroneous predictions~\cite{raissi2018deep,sun2020surrogate}. 

In this study, we aim to tackle this problem from a multi-task learning perspective because it shares several features with PINN in common. In the multi-task learning, the ultimate goal is to train a network on all tasks jointly. Towards this goal, multi-task learning faces the same problem arising from unmatched gradients. More specifically, the gradient might be dominated by the value from one task at the cost of degrading the performance of the other task. In addition to imbalanced gradients, the gradients corresponding to different tasks (or loss terms in PINN) might be conflicting along the direction of descent with one another in a way that is detrimental to the progress of the optimization. These factors combined together result in the fact that the optimizer struggles to make progress in optimizing the weights of the network because the reduction in the loss value specific to one task eventually leads to the oscillation of losses in other tasks (see the demonstration in Section~\ref{sec:examples}).

To resolve this issue,~\citet{yu2020gradient} proposed a projecting conflicting gradients (PCGrad) approach to minimize the gradient interference, which consists in projecting the gradient of a task onto the norm plane of any other task that has a conflicting gradient. In this paper, we adopt the PCGrad method to deconflict gradients during the training of PINN, where we treat each loss term as an individual task in the learning process. Specifically, consider two gradients $\bm{\Delta}_i$ and $\bm{\Delta}_j$ corresponding to the $i$-th and the $j$-th loss term in PINN. PCGrad first checks whether there are conflicts between $\bm{\Delta}_i$ and $\bm{\Delta}_j$ using the cosine similarity defined in Eq. (\ref{eq:similarity}).
\begin{equation}\label{eq:similarity}
\omega \left( {{\bm{\Delta} _i},{\bm{\Delta} _j}} \right) = \frac{{{\bm{\Delta} _i} \bullet {\bm{\Delta} _j}}}{{\left\| {{\bm{\Delta} _i}} \right\|\left\| {{\bm{\Delta} _j}} \right\|}}
\end{equation}
where $\left\|  \bullet  \right\|$ denotes the norm of the corresponding vector. 

The cosine similarity results in a value within the range $\left[ { - 1,1} \right]$, where -1 denotes exactly the opposite direction, 1 means exactly the same, and 0 indicates orthogonality or decorrelation. If the cosine similarity between $\bm{\Delta} _i$ and $\bm{\Delta} _j$ is negative, then PCGrad projects $\bm{\Delta} _i$ to the norm plane of $\bm{\Delta} _j$ or the other way around. If the cosine similarity between $\bm{\Delta} _i$ and $\bm{\Delta} _j$ is non-negative, then the original gradients $\bm{\Delta} _i$ and $\bm{\Delta} _j$ remain the same. Suppose we project $\bm{\Delta} _i$ to the norm plane of $\bm{\Delta} _j$, then we have the gradient of $\bm{\Delta} _i$ after the projection as: 
\begin{equation}
\bm{\Delta} _i^{PC} = {\bm{\Delta} _i} - \frac{{{\bm{\Delta} _i} \bullet {\bm{\Delta} _j}}}{{{{\left\| {{\bm{\Delta} _j}} \right\|}^2}}}{\bm{\Delta} _j}
\end{equation}
where $\bm{\Delta} _i^{PC}$ denotes the gradient after the projection. 

\begin{figure}[!ht]
    \centering
    \includegraphics[scale=0.7]{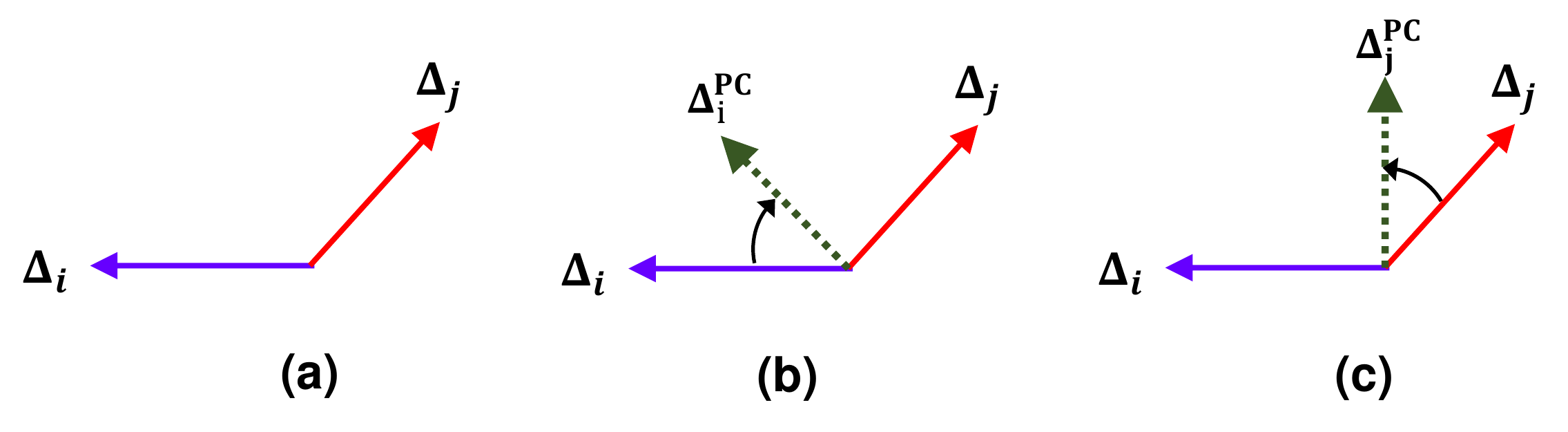}
    \caption{Gradient deconflict process in PCGrad. (a) Two conflicting gradients $\bm{\Delta}_i$ and $\bm{\Delta}_j$. (b) PCGrad projects the gradient $\bm{\Delta}_i$ onto the norm vector of the gradient $\bm{\Delta}_j$. (c) PCGrad projects the gradient $\bm{\Delta}_j$ onto the norm vector of the gradient $\bm{\Delta}_i$}
    \label{fig:conflict_check}
\end{figure}

\begin{algorithm}[!ht]
\SetAlgoLined
\SetNoFillComment
\SetKwFor{For}{for (}{) $\lbrace$}{$\rbrace$}
\caption{Conflicting gradients projection using PCGrad}\label{alg:flowchart}
\KwData{neural network weights $\bm{\theta}$, number of tasks $\mathcal{K}$}

$\bm{\Delta}_i$ $\leftarrow $ ${\nabla _{\bm{\theta}} }{\mathcal{L}_i}\left( {\bm{\theta}}  \right),\;\;\forall i = 1,2, \cdots ,\mathcal{K}$ \tcc{$\mathcal{L}_i$ denotes the $i$-th loss term in PINN}

$\bm{\Delta} _i^{PC} \leftarrow {\bm{\Delta} _i} \; \; \forall i$\

\For{$i = 1;\ i <= \mathcal{K};\ i = i + 1$}{
    \For{$j\mathop  \sim \limits^{\textrm{uniformly}} \left[ {1,2, \cdots ,m, \cdots ,\mathcal{K}} \right],\;$ where $m \ne i$}{
    
      \If{$\omega \left( {{\bm{\Delta} _i^{PC}},{\bm{\Delta} _j}} \right) < 0$}{
        Set $\bm{\Delta} _i^{PC} = \bm{\Delta} _i^{PC} - \frac{{{\bm{\Delta} _i^{PC}} \bullet {\bm{\Delta} _j}}}{{{{\left\| {{\bm{\Delta} _j}} \right\|}^2}}}{\bm{\Delta} _j}$\ \tcc{Subtract the projection of $\bm{\Delta} _i^{PC}$ onto $\bm{\Delta}_j$}
      }
    }
}
\Return update $\bm{\Delta} \bm{\theta}  = \sum\limits_{i = 1}^\mathcal{K} {\bm{\Delta} _i^{PC}} $
\end{algorithm}

Fig. \ref{fig:conflict_check} demonstrates the core idea in PCGrad. As it can be observed in Fig. \ref{fig:conflict_check}(a), there is a high degree of conflict between the two gradients $\bm{\Delta} _i$ and $\bm{\Delta} _j$. PCGrad either projects the gradient $\bm{\Delta} _i$ onto the norm vector of the gradient $\bm{\Delta}_j$ as illustrated in Fig. \ref{fig:conflict_check}(b), or projects the gradient $\bm{\Delta}_j$ onto the norm vector of the gradient $\bm{\Delta}_i$ as shown in Fig. \ref{fig:conflict_check}(c). Such operation amounts to removing the conflicting component from the gradient task, thus mitigating the destructive gradient interference among different tasks. In a similar way, PCGrad repeats the same procedures for all the other tasks following a randomly sampled order. Algorithm~\ref{alg:flowchart} summarizes the steps in PCGrad for projecting conflicting gradients in PINNs. 

As the gradient projection operation accounts for the gradient information for all the tasks in a holistic manner, it significantly mitigates the conflicts among different tasks and results in a set of gradients with minimal gradient interference.

\section{Numerical Examples}\label{sec:examples}
In this section, we demonstrate the proposed framework for MSS reliability assessment using a small-scale MSS of a single propulsion module in a railway system under either time-independent state transitions or time-dependent state transitions. We also illustrate its performance in assessing the reliability of a medium-scale MSS regarding a flow transmission system. The performance of the proposed framework is examined in comparison with the solution derived by the differential equation solver implemented in Matlab.   

\subsection{Example 1}\label{subsec:example_1}

\begin{figure}[!ht]
    \centering
    \includegraphics[scale=0.6]{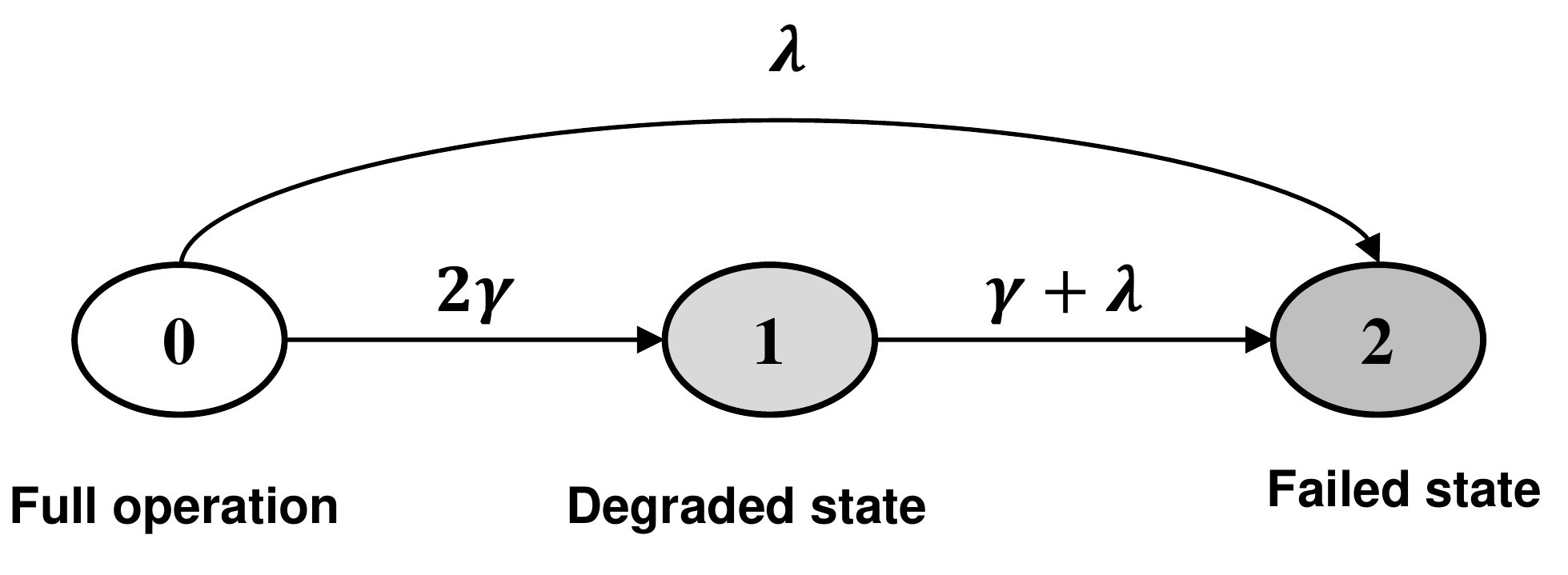}
    \caption{A three-state transition diagram representing the performance deterioration of a propulsion module. The label along each arc denotes the corresponding transition probability from one state to another state. }
    \label{fig:three_state_problem}
\end{figure}

Consider a single propulsion module in a multi-voltage propulsion system designed for the Italian high-speed railway system. The propulsion module consists of a series of four components (transformer, filter, inverter and motor) and two parallel converters~\cite{trivedi2017reliability}. The propulsion module can be represented by a three-state Markov model as illustrated in Fig.~\ref{fig:three_state_problem}. The three states in the propulsion module correspond to three different levels of power delivery as described below:
\begin{enumerate}
    \item Full operation: State 0 denotes a fully operational state. In this state, the propulsion delivers the maximum power (2200 kW) when all components are working.
    
    \item Degraded state: State 1 indicates a degraded state. In this state, only half of the power (1100 kW) is delivered when all the series components, and one converter out of the two works normally. 
    
    \item Failed state: State 2 means the state of complete failure  with no power delivered. 
\end{enumerate}

The transition probabilities across the three states are described by the following matrix:
\begin{equation}\label{eq:transition_rate}
\bm{Q} = \left[ {\begin{array}{*{20}{c}}
{ - \left( {2\gamma  + \lambda } \right)}&{2\gamma}&\lambda \\
0&{ - \left( {\gamma  + \lambda } \right)}&\left( {\gamma  + \lambda } \right)\\
0&0&0
\end{array}} \right]
\end{equation}
where $\gamma$ denotes the sum of failure rates of the series components $\left( \gamma = 2.8 \times 10^{-5} \;\; \text{hour}^{-1} \right)$, and $\lambda$ denotes the failure rate associated with each converter $\left( \lambda= 7.26 \times 10^{-5} \;\; \text{hour}^{-1} \right)$. 

As it can be seen from Eq. (\ref{eq:transition_rate}), the transition rate is time-independent, thus leading to a homogeneous continuous-time Markov chain (CTMC). The propulsion module starts at full capacity and denote its initial state as below:
\begin{equation}
 {\bm{s}_0} = \left[ {1,0,0} \right]   
\end{equation}

In MSS, we are interested in estimating the probabilities associated with the three states in the operation time horizon $t \in \left( {0,T} \right]$. Following the methodology described in Section~\ref{sec:proposed_methodology}, together with Eq. (\ref{eq:transition_equation}), we derive four loss terms that belong to two individual groups in the PINN as formulated below:
\begin{equation}
\begin{array}{l}
{\mathcal{L}_u}\left( \bm{\theta}  \right) = {\left[ {{u_{NN}}\left( {t = 0;\bm{\theta} } \right) - {\bm{s}_0}} \right]^2}\\
{\mathcal{L}_f}\left( \bm{\theta}  \right) = \left\{ \begin{array}{l}
{\left\{ {\frac{{du_{NN}^0\left( {t;\bm{\theta} } \right)}}{{dt}} - u_{NN}^0\left( {t;\bm{\theta} } \right) \times \left[ { - \left( {2\gamma  + \lambda } \right)} \right]} \right\}^2}\\
{\left\{ {\frac{{du_{NN}^1\left( {t;\bm{\theta} } \right)}}{{dt}} - \left[ {u_{NN}^0\left( {t;\bm{\theta} } \right) \times 2\gamma  - u_{NN}^1\left( {t;\theta } \right) \times \left( {\gamma  + \lambda } \right)} \right]} \right\}^2}\\
{\left\{ {\frac{{du_{NN}^2\left( {t;\bm{\theta} } \right)}}{{dt}} - \left[ {u_{NN}^0\left( {t;\bm{\theta}} \right) \times \lambda  + u_{NN}^1\left( {t;\bm{\theta} } \right) \times \left( {\gamma  + \lambda } \right)} \right]} \right\}^2}
\end{array} \right.
\end{array}
\end{equation}
where $u_{NN}^i \left(t;\bm{\theta} \right)$ denotes the prediction of neural network on the $i$-th state in the three-state MSS at the time instant $t$.

\begin{figure}[!ht]
    \centering
    \includegraphics[scale=0.418]{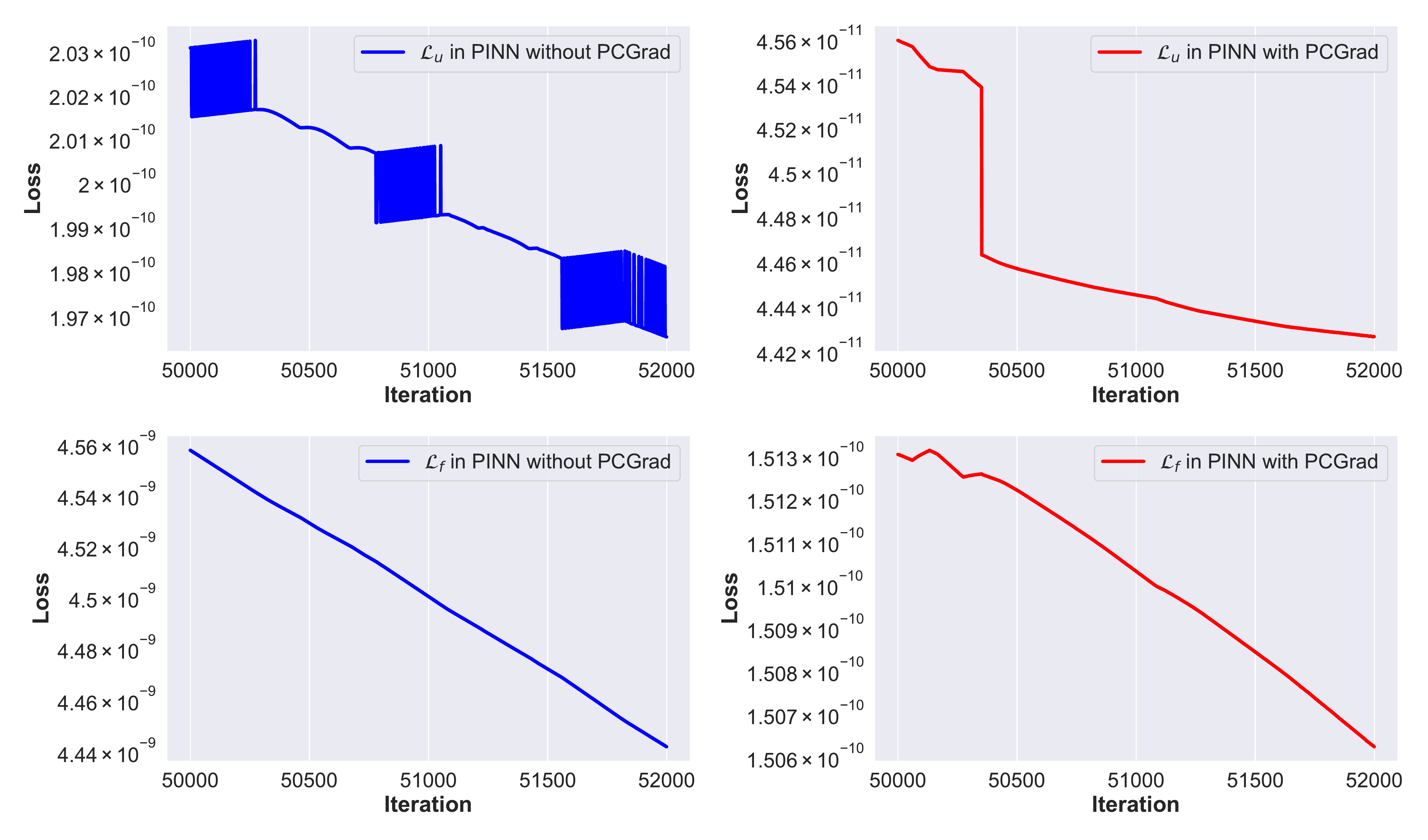}
    \caption{A snapshot of the convergence of the two loss groups $\mathcal{L}_u$ and $\mathcal{L}_f$}
    \label{fig:loss_convergence}
\end{figure}

Ideally, the value of each component in the loss group $\mathcal{\bm{L}}_f$ should be strictly zero. In PINN, we approximate these equations by embedding a soft loss term in the objective function. In this example, the PINN consists of two hidden layers with each hidden layer having 50 units and the Tanh activation function. The last layer of the PINN is a fully-connected layer, and it has three outputs with each output corresponding to one state in the MSS, and softmax is employed as the activation function in the fully-connected layer to ensure that the range of output value is within the range $\left[ {0,1} \right]$\footnote[1]{In this paper, PINN has the same architecture across the three numerical examples. The only difference is the number of outputs in the fully-connected layer.}. To train the neural network, we generate 5000 collocation points for $t \in \left[ {0, 60,000} \right]$ with equal intervals representing Eq. (\ref{eq:transition_equation}). Next, we use the Adam algorithm with a learning rate of 0.001 to optimize the weights of the neural network. The two PINNs trained using Adam with and without PCGrad have the same architecture and initial weights. In the PINN with PCGrad, the two loss groups $\mathcal{L}_u$ and $\mathcal{L}_f$ are treated as two individual tasks in the PINN. Whereas, in the PINN without PCGrad, the two loss groups are combined together via equal weights\footnote[2]{In the case of no PCGrad, the losses are combined following the same way in the subsequent two examples.}. The paradigm of adopting the proposed PINN-based framework to model MSS reliability assessment apparently differs from existing methods for MSS reliability analysis. The proposed PINN-based framework provides an insightful point of view to take advantage of the power in neural network to tackle this challenging problem.

Fig. \ref{fig:loss_convergence} illustrates a snapshot demonstrating the convergence of the two loss groups $\mathcal{L}_u$ and $\mathcal{L}_f$ during PINN training. Obviously, without PCGrad, the decrease in the loss group $\mathcal{L}_f$ (note loss $\mathcal{L}_f$ dominates in the loss function in the snapshot) leads to severe oscillation of the other loss group $\mathcal{L}_u$, which eventually translates into the slow convergence of the PINN. In contrast, the situation is completely different after PCGrad is applied. The values of both loss groups $\mathcal{L}_u$ and $\mathcal{L}_f$ drop down in a steady trend, and PCGrad converges to a solution with a loss value that is much lower than that of PINN without PCGrad by one order of magnitude. 

\begin{figure}[!ht]
    \centering
    \includegraphics[scale=0.48]{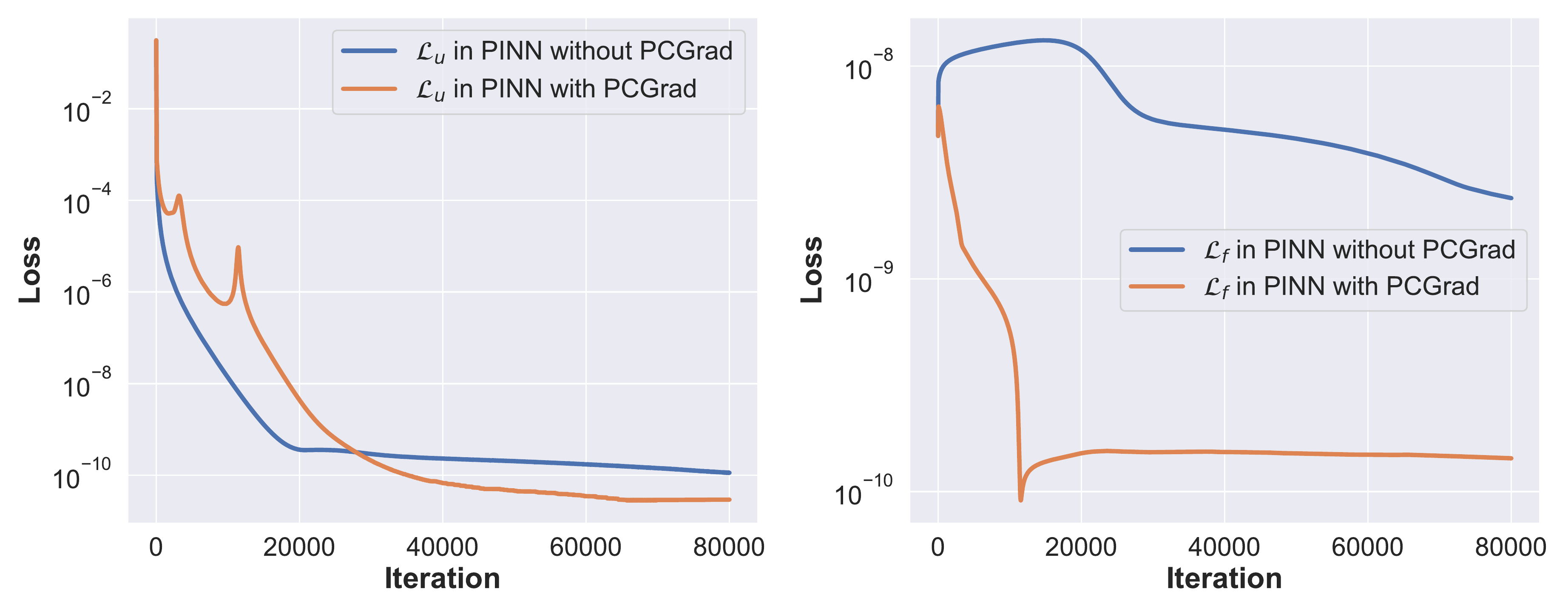}
    \caption{Convergence performance comparison of two loss groups in the MSS for the homogeneous propulsion module}
    \label{fig:performance_comparison}
\end{figure}

Fig. \ref{fig:performance_comparison} compares the convergence performance of the two loss groups after the PINN is trained for 80,000 iterations. It can be noted that both $\mathcal{L}_u$ and $\mathcal{L}_f$ converge to a much lower values in the PINN with PCGrad than that of PINN without PCGrad. In particular, PINN with PCGrad achieves much better performance in approximating the transition equations $\mathcal{L}_f$ than the case of no PCGrad as partially reflected by the significant gap in the early stage of iterations as indicated in Fig. \ref{fig:performance_comparison}. To examine the solution quality of PINN trained with PCGrad, we derive solutions to the PDE using the Runge–Kutta method for the 5000 time instants within the time range $\left [0, 80,000\right]$ with a step size of 16 in Matlab. Next, we use the root squared mean error (RMSE)  in relation to the results from Runge-Kutta using Matlab solver that is averaged over the three states as the performance metric to compare the probabilities associated with the three states generated by PINN with that of the Runge–Kutta method. Note that PINN is trained with the data in the time range $\left [ 0, 60000\right]$. In other words, we examine the performance of PINN in both interpolating and extrapolating the state probabilities. 

\begin{figure}[!ht]
    \centering
    \includegraphics[scale=0.5]{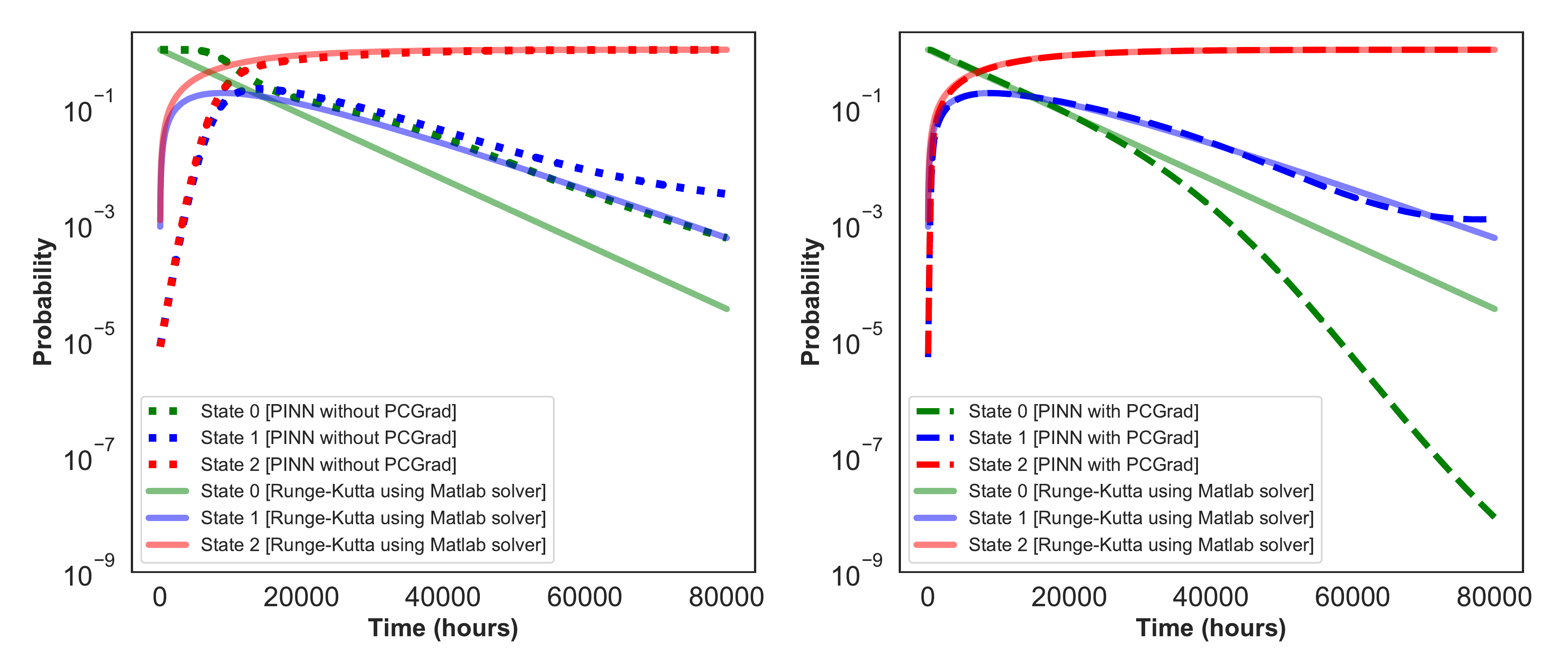}
    \caption{Performance comparison of PINN's predictions on the probabilities associated with each state over time in the MSS}
    \label{fig:performance_comparison_example_1}
\end{figure}

\begin{figure}[!ht]
    \centering
    \includegraphics[scale=0.5]{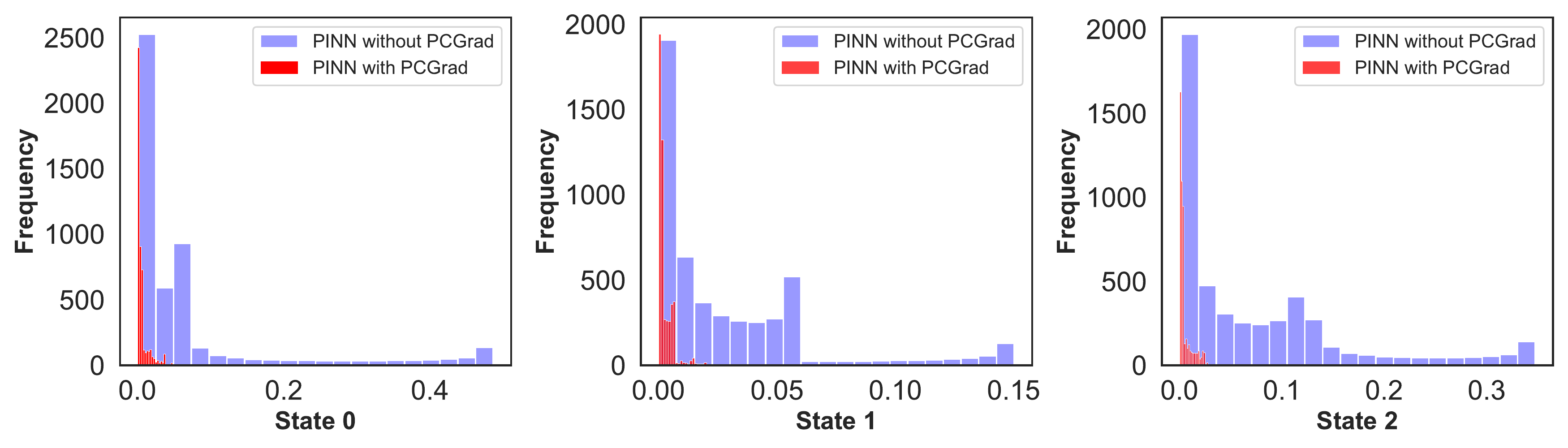}
    \caption{Histogram of the mean absolute error between PINN and the Matlab solver with respect to each system state in the homogeneous propulsion system}
    \label{fig:histogram_small_homo}
\end{figure}

\begin{table}[!ht]
    \centering
    \caption{Performance comparison of PINN with and without PCGrad in the reliability assessment of the homogeneous flow transmission system}
    \begin{tabular}{c|c|c|c}
    \toprule
     \textbf{Methods}    &  \multicolumn{3}{c}{\textbf{Time Range}}\\
    \midrule
    & $t \in \left(0, 80000\right]$ & $t \in \left(0, 60000\right]$ & $t \in \left(60000, 80000\right]$\\
    \hline
     RMSE of PINN without PCGrad    & 0.111139 & 0.1461463 & 0.0062300\\
     RMSE of PINN with PCGrad       & 0.008277 & 0.0108075 & 0.0006953\\
    \bottomrule
    \end{tabular}
    \label{tab:performance_comparison_example_1}
\end{table}

Fig. \ref{fig:performance_comparison_example_1} compares the predictions of the two PINN models with the Matlab solver. It can be observed that PCGrad outperforms the optimization without PCGrad when estimating the probabilities of state 1 and state 2. At the same time, PCGrad overestimates the probability associated with state 0. Another interesting observation is that both optimization methods with and without PCGrad perform well in extrapolation for the time range $t \in \left(60,000, 80,000 \right]$ due to the incorporation of PDE equations. Fig.~\ref{fig:histogram_small_homo} displays the histogram of the mean absolute error between PINN's predictions and the Matlab solver with respect to the three system states. Not surprisingly, PINN with PCGrad achieves substantially lower mean absolute error in all three system states than that of PINN without PCGrad. In other words, PCGrad significantly improves the quality of the state estimation in MSS. Table~\ref{tab:performance_comparison_example_1} summarizes the performance comparison quantitatively. Clearly, PINN with PCGrad achieves a RMSE that is lower than that of PINN without PCGrad by at least one order of magnitude across the three time ranges.

\subsection{Example 2}
In this example, we extend the Example~\ref{subsec:example_1} by imposing time-inhomogeneity in the component failures. 
Differing from Example~\ref{subsec:example_1}, the transition rate is time-dependent in this example, thus leading to a non-homogeneous CTMC. In particular, we assume that the transition rates follow a Weibull distribution and the corresponding transition rates are defined as below:
\begin{equation}
\begin{array}l
\lambda\left( t \right) = \lambda_0\alpha t^{\alpha-1}  \\
\gamma\left( t \right) = \gamma_0\beta t^{\beta-1}     \\
\end{array}
\end{equation}
where $\lambda_0$ and $\gamma_0$ denote the initial failure rates associated with each state, and their values are the same as the Example in Section~\ref{subsec:example_1}. More specifically, $\lambda_0$ has a value of $7.26 \times 10^{-5} \;\; \text{hour}^{-1}$ and $\gamma_0$ has a value of $2.8 \times 10^{-5} \;\; \text{hour}^{-1}$. 

For the sake of illustration, we set the parameters in the Weibull distribution as $\alpha = 2.0$ and $\beta = 2.0$. The corresponding transition matrix of this problem is shown in Eq. (\ref{eq:transition_rate_non_home}).

\begin{equation}\label{eq:transition_rate_non_home}
\bm{Q}\left( t \right) = \left[ {\begin{array}{*{20}{c}}
{ - \left( {2\gamma\left( t \right)  + \lambda\left( t \right) } \right)}&{2\gamma\left( t \right)}&\lambda\left( t \right) \\
0&{ - \left( {\gamma\left( t \right)  + \lambda\left( t \right) } \right)}&\left( {\gamma\left( t \right)  + \lambda\left( t \right) } \right)\\
0&0&0
\end{array}} \right]
\end{equation}

\begin{figure}[!ht]
    \centering
    \includegraphics[scale=0.48]{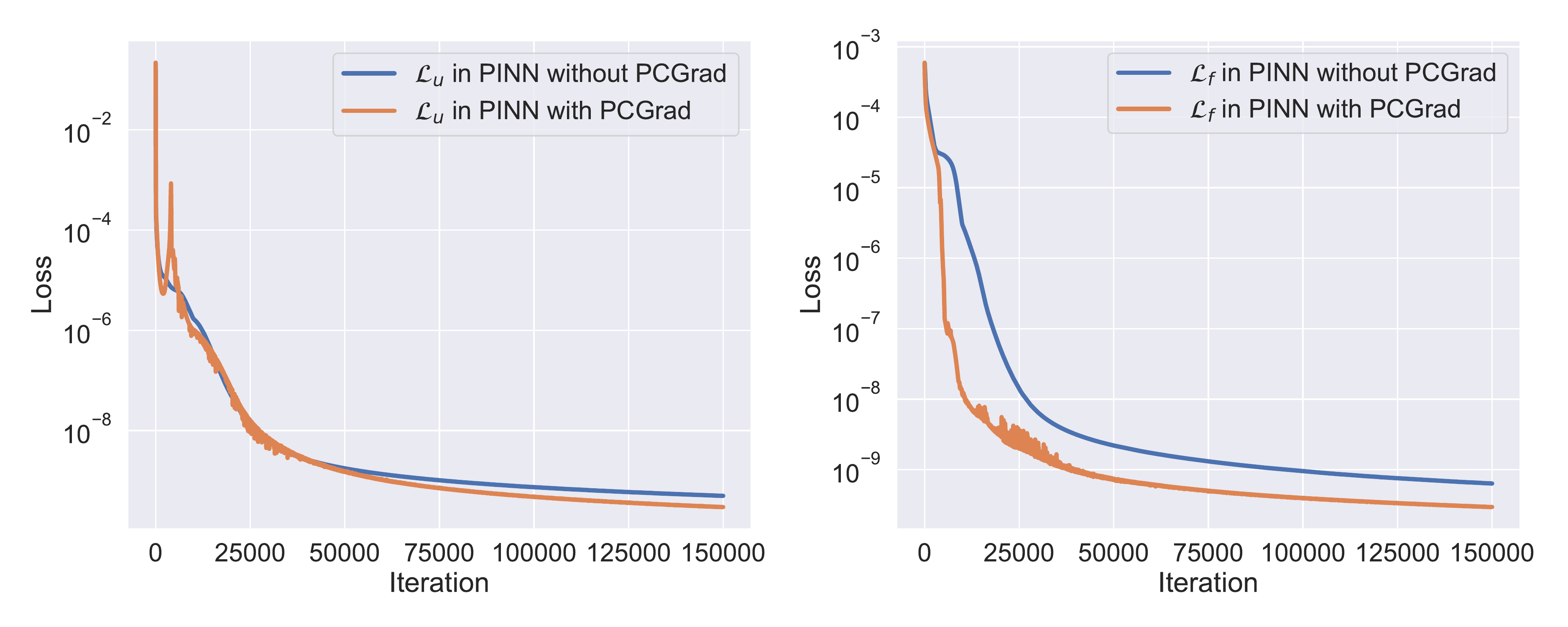}
    \caption{Convergence performance comparison of PINN for reliability assessment of the non-homogeneous propulsion module}
    \label{fig:performance_comparison_non_homo}
\end{figure}

In the non-homogeneous CTMC, one difference worthy of mention is that each loss term in the loss group $\mathcal{L}_f$ now involves the input time $t$ when constructing the ODE residual. Following similar steps, the problem can be formulated using the proposed PINN-based framework. To train the PINN, we generate 300 collocation points within the time range $\left [0, 300\right]$ with equal intervals. The initial state is the same as Example \ref{sec:examples}. The architecture of PINN is exactly the same as in Example~\ref{sec:examples}. As the gradient during back-propagation is complex in the non-homogeneous CTMC case, we impose a monotonically decreasing learning rate that follows a polynomial decay schedule with an initial learning rate of $1*10^{-3}$ and a final learning rate of $8*10^{-5}$. Fig. \ref{fig:performance_comparison_non_homo} compares the convergence performance of the two loss groups during the 150000 iterations. Clearly, PINN with PCGrad converges to lower loss values in both loss groups $\mathcal{L}_u$ and $\mathcal{L}_f$ than that of the PINN without PCGrad. 

\begin{figure}[!ht]
    \centering
    \includegraphics[scale=0.51]{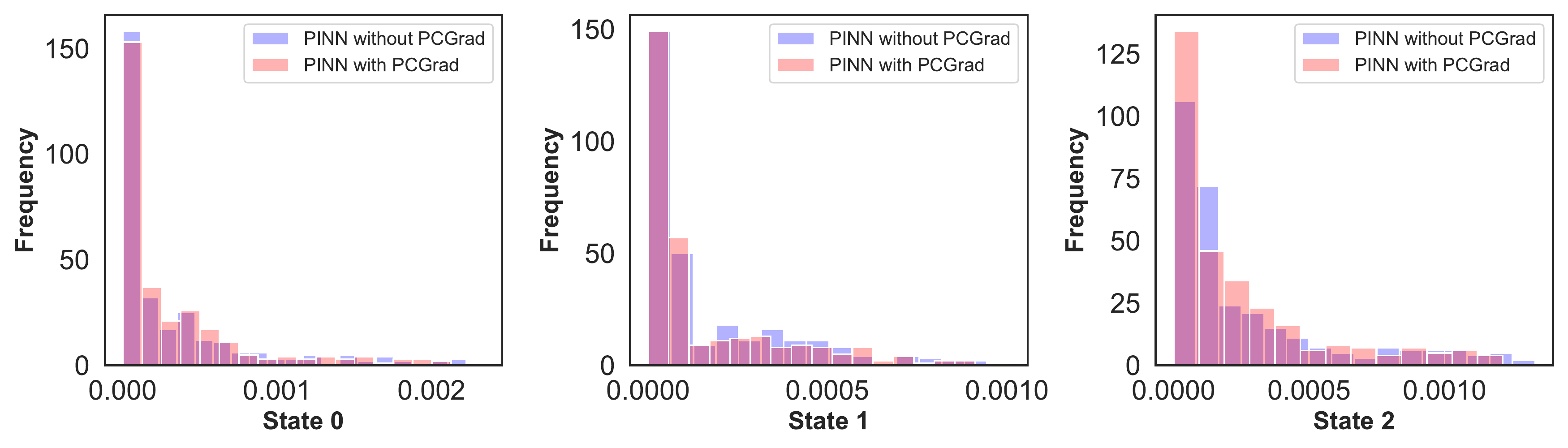}
    \caption{Histogram of the mean absolute error between PINN and the Matlab solver with respect to each system state in the non-homogeneous propulsion module}
    \label{fig:histogram_small_non_homo}
\end{figure}

To test the performance of the trained PINN model, we generate another 301 points within the time range $\left [0, 300\right]$ with equal intervals. Fig.~\ref{fig:histogram_small_non_homo} illustrates the histogram of the mean absolute error between PINN's predictions with Matlab's solutions with respect to the three system states. As it can be observed, PINN with PCGrad has similar performance when estimating the probability corresponding to state 0, while it achieves slightly better performance than the case of no PCGrad when predicting the probabilities associated with state 1 and state 2. 

\begin{figure}[!ht]
    \centering
    \includegraphics[scale=0.51]{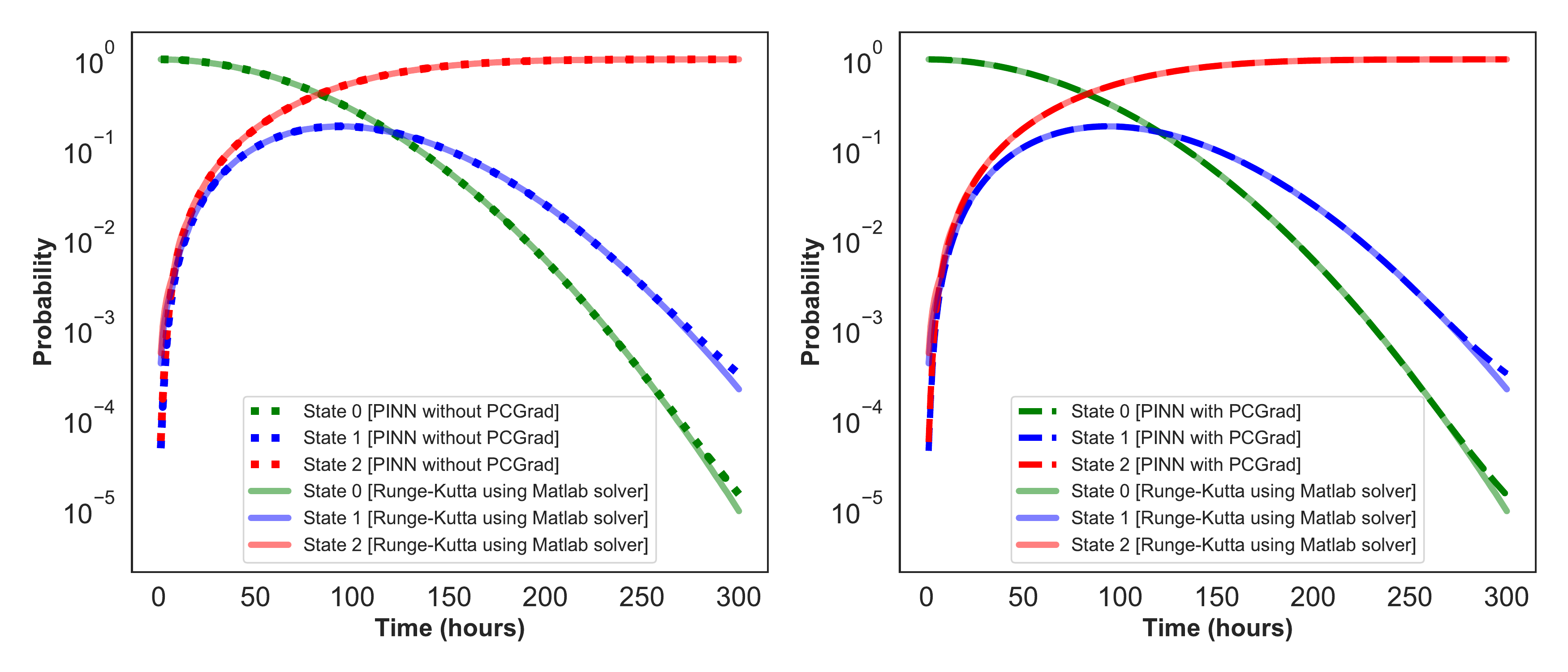}
    \caption{Performance comparison of PINN's predictions on the probabilities associated with each state over time in the non-homogeneous propulsion system}
    \label{fig:comparison_non_homo}
\end{figure}

Next, we evaluate the quality of the solutions of PINN quantitatively by comparing them with solutions derived by the Matlab solver, as shown in Fig.~\ref{fig:comparison_non_homo}. As it can be seen, PINN accurately captures the changing trend of probability associated with each system state over time. Table \ref{tab:performance_comparison_non_homo} summarizes the RMSE between the solutions of PINN with and without PCGrad and the Matlab solutions. PINN with PCGrad achieves a 8.86\% of reduction in RMSE in comparison with the RMSE of PINN without PCGrad.

\begin{table}[!ht]
    \centering
    \caption{Performance comparisons of PINN with and without PCGrad in the reliability assessment of the non-homogeneous flow transmission system}
    \begin{tabular}{c|c}
    \toprule
     \textbf{Methods}    &  \textbf{Time Range}\\
    \midrule
    & $t \in \left[0, 300\right]$ \\
    \hline
     RMSE of PINN without PCGrad    & 0.00048846 \\
     RMSE of PINN with PCGrad       & 0.00044518 \\
    \bottomrule
    \end{tabular}
    \label{tab:performance_comparison_non_homo}
\end{table}

\subsection{Example 3}

\begin{figure}[!ht]
    \centering
    \includegraphics[scale=0.46]{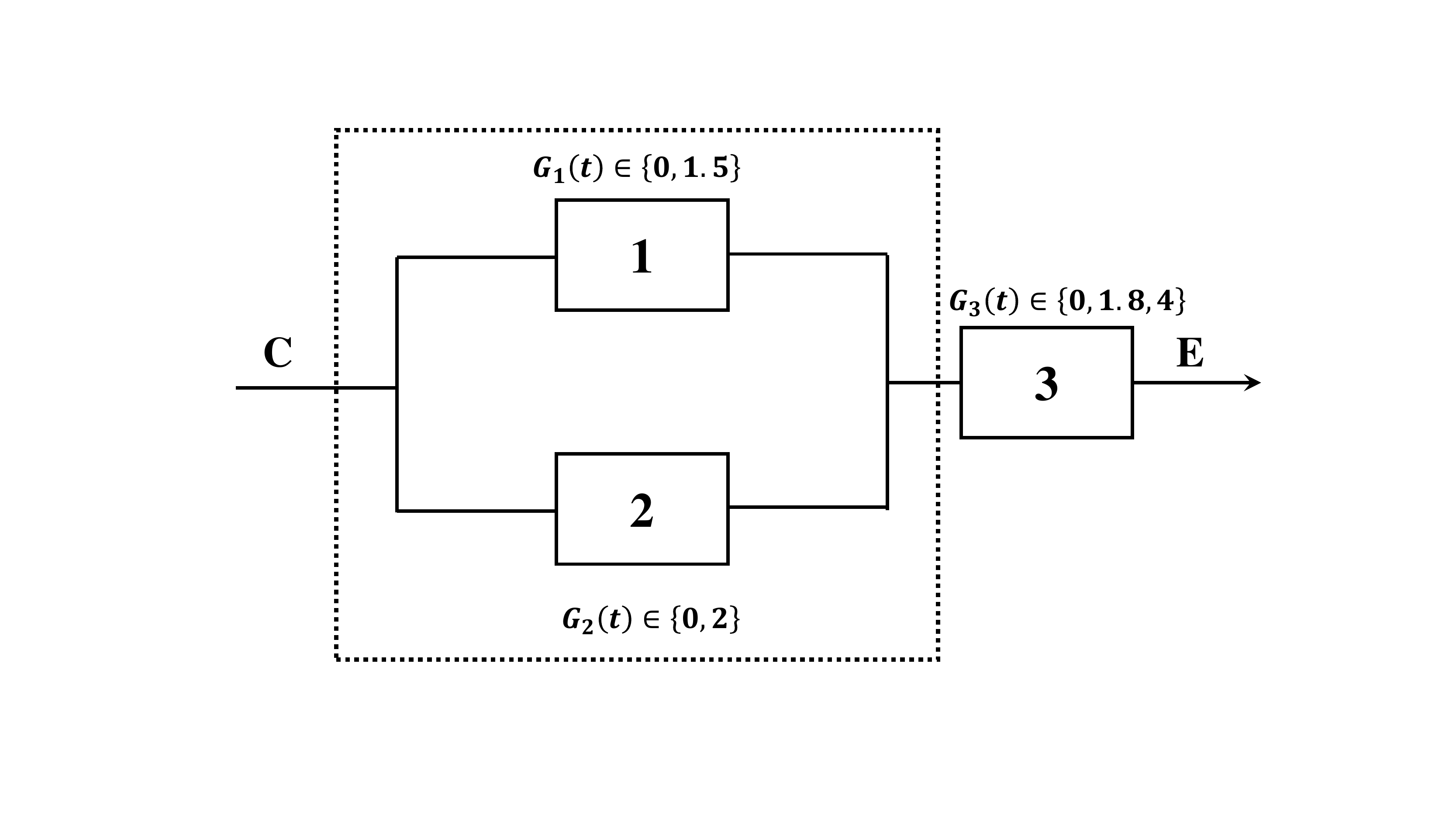}
    \caption{MSS-based representation of the flow transmission system}
    \label{fig:large_example_MSS_structure}
\end{figure}

Consider a flow transmission system consisting of three pipes. Fig.~\ref{fig:large_example_MSS_structure} shows the MSS structure of the flow transmission system~\cite{lisnianski2010multi}, where the oil flows from point C to point E in the flow transmission system. The performance of the flow transmission system is measured by its capacity in the unit of tons per minute. Both element 1 and element 2 have two states: operational and failed state. In the operational state, element 1 and element 2 have a capacity of 1.5 and 2 tons per minute, respectively. Whereas, their capacity degrades to zero if they are in a state of total failure. Differing from element 1 and element 2, element 3 has three states: a state of total failure corresponding to a capacity of 0, a state of partial failure corresponding to a capacity of 1.8 tons per minute, and a fully operational state with a capacity of 4 tons per minute. 

\begin{figure}[!ht]
    \centering
    \includegraphics[scale=0.42]{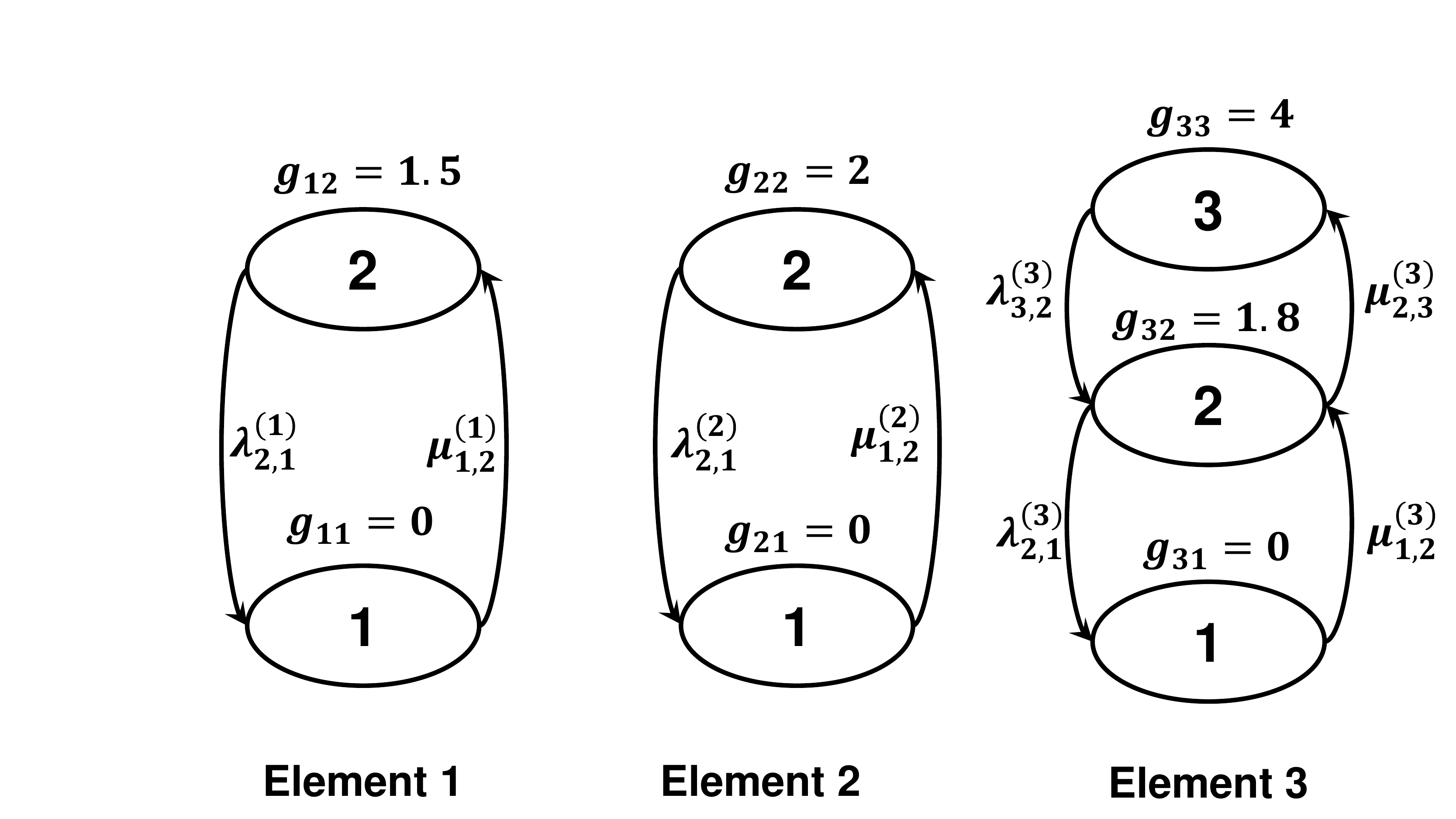}
    \caption{Component-wise state transitions in the flow transmission system}
    \label{fig:large_example_component_state_transitions}
\end{figure}

Fig. \ref{fig:large_example_component_state_transitions} illustrates the state transitions of each component in the flow transmission system, where $\lambda_{i,j}^{(k)}$ and $\mu_{j,i}^{(k)}$ denote the failure rate and repair rate associated with the $k$-th element when the element transitions between state $i$ and state $j$ in the flow transmission system. The specific values of the failure rates and repair rates are shown as below:
\begin{equation}
\begin{array}{l}
\lambda_{2,1}^{(1)} = 7 \;\; \text{year}^{-1}, \mu_{1,2}^{(1)} = 100 \;\; \text{year}^{-1};\\
\lambda_{2,1}^{(2)} = 10 \;\; \text{year}^{-1}, \mu_{1,2}^{(2)} = 80 \;\; \text{year}^{-1};\\
\lambda_{3,2}^{(3)} = 10 \;\; \text{year}^{-1}, \mu_{2,3}^{(3)} = 110 \;\; \text{year}^{-1}; \\
\lambda_{2,1}^{(3)} = 7 \;\; \text{year}^{-1}, \mu_{1,2}^{(3)} = 120 \;\; \text{year}^{-1}. \\
\end{array}
\end{equation}

\begin{figure}[!ht]
    \centering
    \includegraphics[scale=0.64]{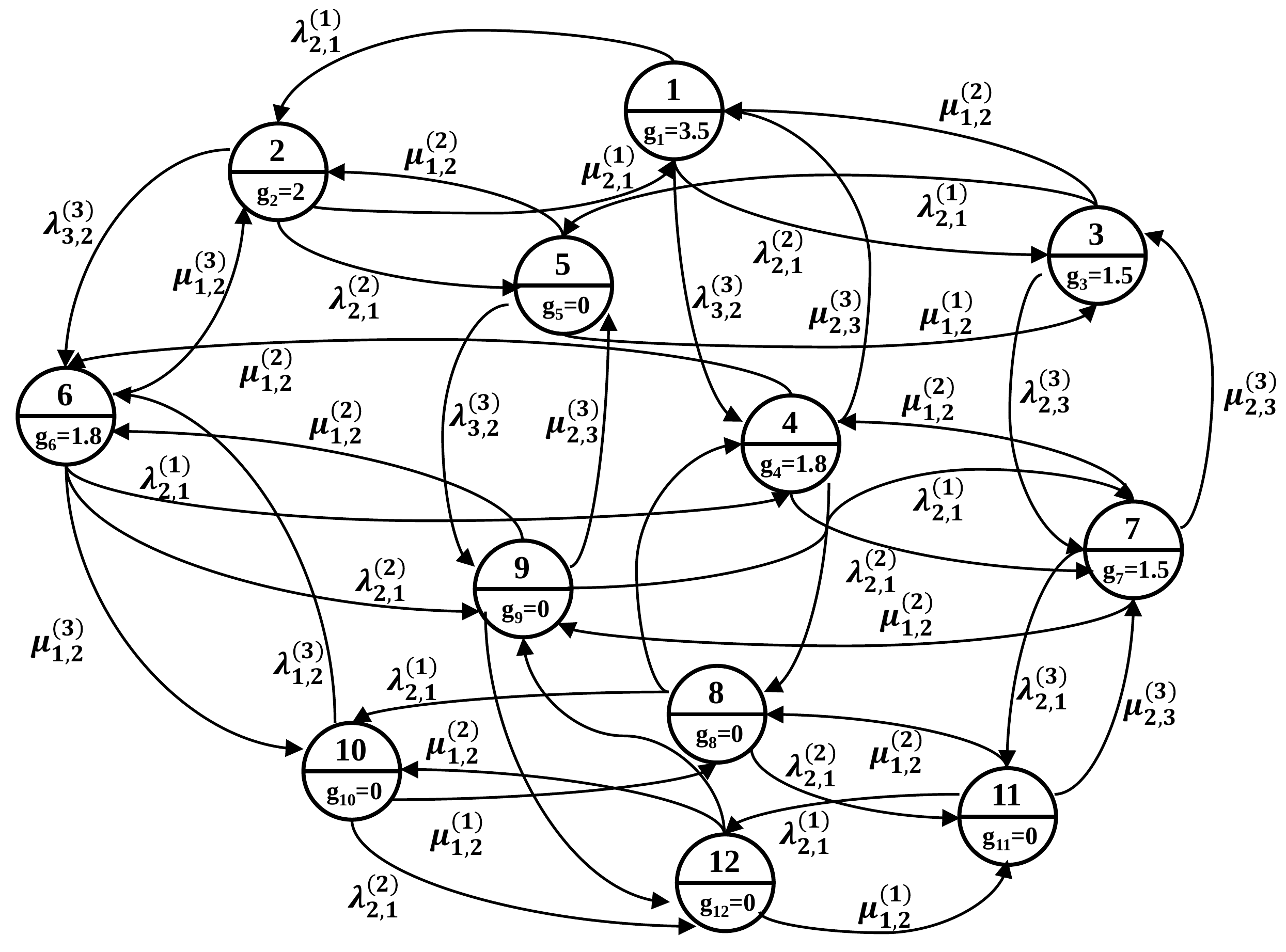}
    \caption{State transition diagram at the system level of the flow transmission}
    \label{fig:large_example}
\end{figure}

In this application, we are interested in the system-level performance, which is measured by the maximum flow that can be transmitted from point C to point E. At the system level, there are 12 states ($2 \times 2 \times 3$) in total. The state transition diagram at the system level is shown in Fig. \ref{fig:large_example}, where the corresponding system performance is presented in the lower parts of the circle, and the label along each arc denotes the transition probability from one state to another state. For the details in deriving the system-level state transition diagram, refer to page 83 in chapter 2 of the Ref. \cite{lisnianski2010multi}. 

The differential equations governing the transitions among different system performance rates $p_i\left( t \right) \;\; $ are shown in Eq.~(\ref{eq:transition_eqs}). 
\begin{equation}\label{eq:transition_eqs}
\left\{ \begin{array}{l}
\frac{{d{p_1}\left( t \right)}}{{dt}} = -\left( \lambda_{2,1}^{(1)} + \lambda_{2,1}^{(2)} + \lambda_{3,2}^{(3)}\right)p_1\left( t \right) + \mu_{1,2}^{(1)}p_2\left( t \right) + \mu_{1,2}^{(2)}p_3\left( t \right) + \mu_{2,3}^{(3)}p_4\left( t \right),\\
\frac{{d{p_2}\left( t \right)}}{{dt}} = \lambda_{2,1}^{(1)}p_1\left( t \right) - \left (\mu_{1,2}^{(1)} + \lambda_{2,1}^{(2)} + \lambda_{3,2}^{(3)}\right)p_2\left( t \right) + \mu_{1,2}^{(2)}p_5\left( t \right) + \mu_{2,3}^{(3)}p_6\left( t \right),\\
\frac{{d{p_3}\left( t \right)}}{{dt}} = \lambda_{2,1}^{(2)}p_1\left( t \right) - \left (\mu_{1,2}^{(2)} + \lambda_{2,1}^{(1)} + \lambda_{3,2}^{(3)}\right)p_3\left( t \right) + \mu_{1,2}^{(1)}p_5\left( t \right) + \mu_{2,3}^{(3)}p_7\left( t \right),\\
\frac{{d{p_4}\left( t \right)}}{{dt}} = \lambda_{3,2}^{(3)}p_1\left( t \right) - \left (\mu_{2,3}^{(3)} + \lambda_{2,1}^{(1)} + \lambda_{2,1}^{(2)} + \lambda_{2,1}^{(3)}\right)p_4\left( t \right) + \mu_{1,2}^{(1)}p_6\left( t \right) + \mu_{1,2}^{(2)}p_7\left( t \right)+ \mu_{1,2}^{(3)}p_8\left( t \right),\\
\frac{{d{p_5}\left( t \right)}}{{dt}} = \lambda_{2,1}^{(2)}p_2\left( t \right)+\lambda_{2,1}^{(1)}p_3\left( t \right) - \left( \mu_{1,2}^{(2)} + \mu_{1,2}^{(1)} + \mu_{3,2}^{(3)}\right)p_5\left( t \right) + \mu_{2,3}^{(3)}p_9\left( t \right),\\
\frac{{d{p_6}\left( t \right)}}{{dt}} = \lambda_{3,2}^{(3)}p_2\left( t \right)+\lambda_{2,1}^{(1)}p_4\left( t \right) - \left( \mu_{2,3}^{(3)} + \mu_{1,2}^{(1)} + \mu_{2,1}^{(2)}+ \mu_{2,1}^{(3)}\right)p_6\left( t \right) + \mu_{1,2}^{(2)}p_9\left( t \right) + \mu_{1,2}^{(3)}p_{10}\left( t \right),\\
\frac{{d{p_7}\left( t \right)}}{{dt}} = \lambda_{3,2}^{(3)}p_3\left( t \right)+\lambda_{2,1}^{(2)}p_4\left( t \right) - \left( \mu_{2,3}^{(3)} + \mu_{1,2}^{(2)} + \mu_{2,1}^{(1)}+ \mu_{2,1}^{(3)}\right)p_7\left( t \right) + \mu_{1,2}^{(1)}p_9\left( t \right) + \mu_{2,3}^{(3)}p_{11}\left( t \right),\\
\frac{{d{p_8}\left( t \right)}}{{dt}} = \lambda_{2,1}^{(3)}p_4\left( t \right) - \left( \mu_{1,2}^{(3)} + \lambda_{2,1}^{(1)} + \lambda_{2,1}^{(2)}\right)p_8\left( t \right) + \mu_{1,2}^{(1)}p_{10}\left( t \right) + \mu_{1,2}^{(2)}p_{11}\left( t \right),\\
\frac{{d{p_9}\left( t \right)}}{{dt}} =  \lambda_{3,2}^{(3)}p_5\left( t \right) + \lambda_{2,1}^{(2)}p_6\left( t \right) + \lambda_{2,1}^{(1)}p_7\left( t \right) - \left ( \mu_{2,3}^{(3)} + \mu_{1,2}^{(2)} + \mu_{1,2}^{(1)} + \lambda_{2,1}^{(3)} \right) p_9\left( t \right) + \mu_{1,2}^{(3)}p_{12}\left( t \right),\\
\frac{{d{p_{10}}\left( t \right)}}{{dt}} =  \lambda_{2,1}^{(3)}p_6\left( t \right) + \lambda_{2,1}^{(1)}p_8\left( t \right) - \left ( \mu_{1,2}^{(3)} + \mu_{1,2}^{(1)} + \lambda_{1,2}^{(2)} \right) p_{10}\left( t \right) + \mu_{1,2}^{(2)}p_{12}\left( t \right),\\
\frac{{d{p_{11}}\left( t \right)}}{{dt}} =  \lambda_{2,1}^{(3)}p_7\left( t \right) + \lambda_{2,1}^{(2)}p_8\left( t \right) - \left ( \mu_{1,2}^{(3)} + \mu_{1,2}^{(2)} + \lambda_{2,1}^{(1)} \right) p_{11}\left( t \right) + \mu_{1,2}^{(1)}p_{12}\left( t \right),\\
\frac{{d{p_{12}}\left( t \right)}}{{dt}} =  \lambda_{2,1}^{(3)}p_9\left( t \right) + \lambda_{2,1}^{(2)}p_{10}\left( t \right) + \lambda_{2,1}^{(1)}p_{11}\left( t \right) - \left ( \mu_{1,2}^{(3)} + \mu_{1,2}^{(2)} + \mu_{1,2}^{(1)} \right) p_{12}\left( t \right).\\
\end{array} \right. 
\end{equation}

From the state transition diagram in Fig. \ref{fig:large_example}, we observe that there are five unique performance rates at the system level, namely: state 1: $g_1$ = 3.5; state 2: $g_2$ = 2.0; states 4 and 6: $g_4$ = $g_6$ =1.8; states 3 and 7: $g_3$ = $g_7$ =1.5; states 5, 8, 9, 10, 11 and 12: $g_5$ = $g_8$ = $g_9$ = $g_{10}$ = $g_{11}$ = $g_{12}$ = 0. Hence, the reliability of the system-level performance rate is formulated as follows:
\begin{equation}
\begin{array}{l}
    P\left(G=3.5 \right) = p_1\left( t \right);\\
    P\left(G=2.0 \right) = p_2\left( t \right);\\
    P\left(G=1.8 \right) = p_4\left( t \right) + p_6\left( t \right);\\
    P\left(G=1.5 \right) = p_3\left( t \right) + p_7\left( t \right);\\
    P\left(G=0 \right) = p_5\left( t \right) + p_8\left( t \right) + p_9\left( t \right) + p_{10}\left( t \right) + p_{11}\left( t \right) + p_{12}\left( t \right).\\
\end{array}
\end{equation}

Following the framework proposed in Section~\ref{sec:proposed_methodology}, we reformulate this problem in the context of PINN. Specifically,we generate 500 collocation points within the time range $\left[0, 0.2\right]$ with equal intervals, and initialize the state of the system at the time instant $0$ as follows:
\begin{equation}
 {\bm{s}_0} = \left[ {1,0,0,0,0,0,0,0,0,0,0,0} \right]   
\end{equation}

\begin{figure}[!ht]
    \centering
    \includegraphics[scale=0.51]{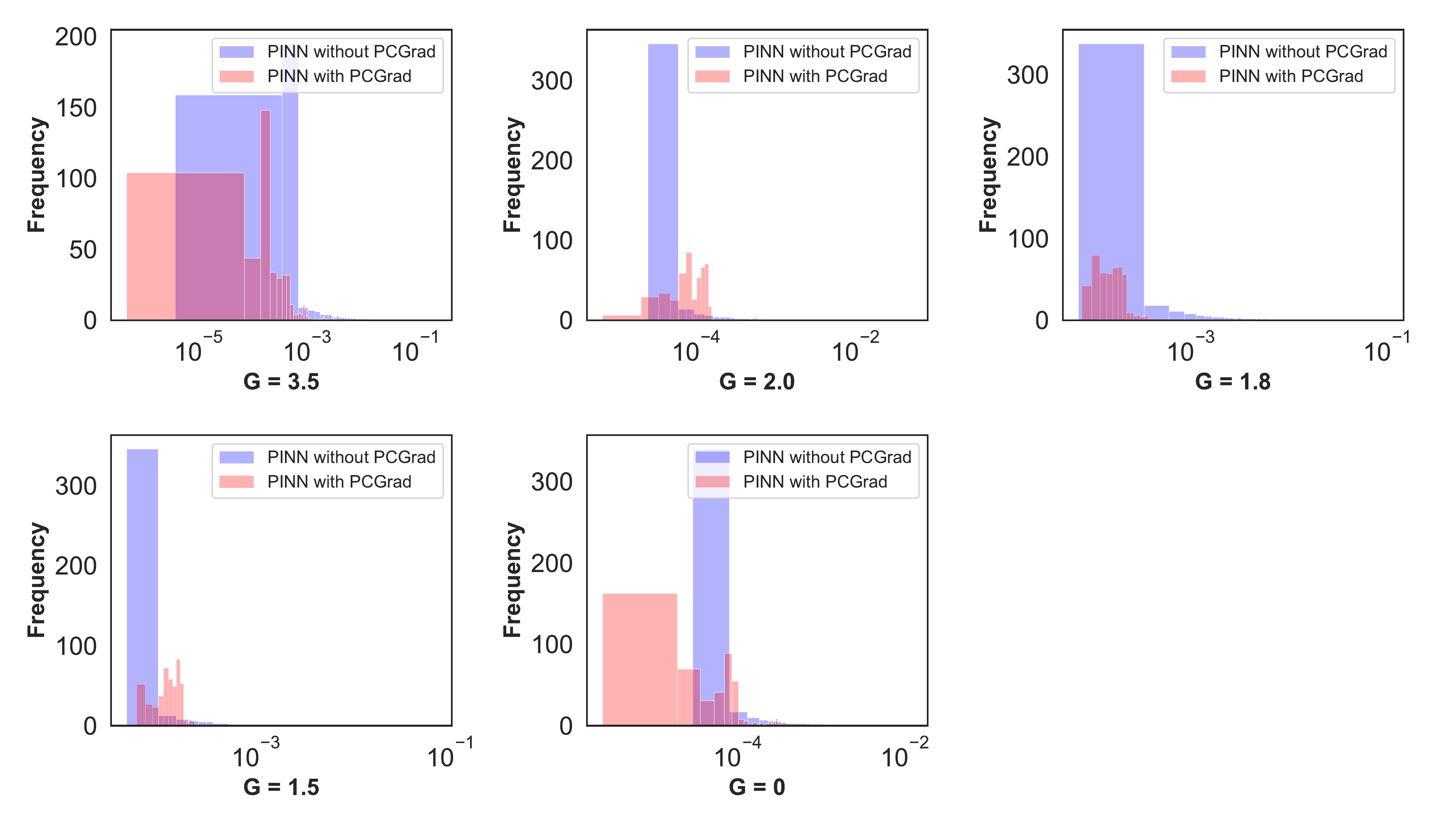}
    \caption{Histogram of the mean absolute error between PINN and the Matlab solver with respect to each system state in the flow transmission system}
    \label{fig:histogram_medium_homo}
\end{figure}

In terms of the architecture of the neural network, the PINN used for reliability assessment in this numerical example has the same configuration as in Example \ref{subsec:example_1}, such as network architecture, learning rate, activation function and number of training steps. The PINN is trained for 40000 iterations using the Adam algorithm with a learning rate of $1*10^{-3}$. To compare the model performance, we generate 501 points within the time range $\left[ 0, 0.2 \right]$ with an equal step size of 0.0004. Fig. \ref{fig:histogram_medium_homo} shows the histogram of the mean absolute error between PINN's predictions and Matlab's solutions regarding the five system-level performance rates. PINN with PCGrad outperforms the case of no PCGrad when system-level performance rates are 3.5, 1.8, and 0. When system-level performance rates $G$ are at 2.0 and 1.5, PCGrad maintains almost the same level of performance as the case of no PCGrad.

\begin{figure}[!ht]
    \centering
    \includegraphics[scale=0.51]{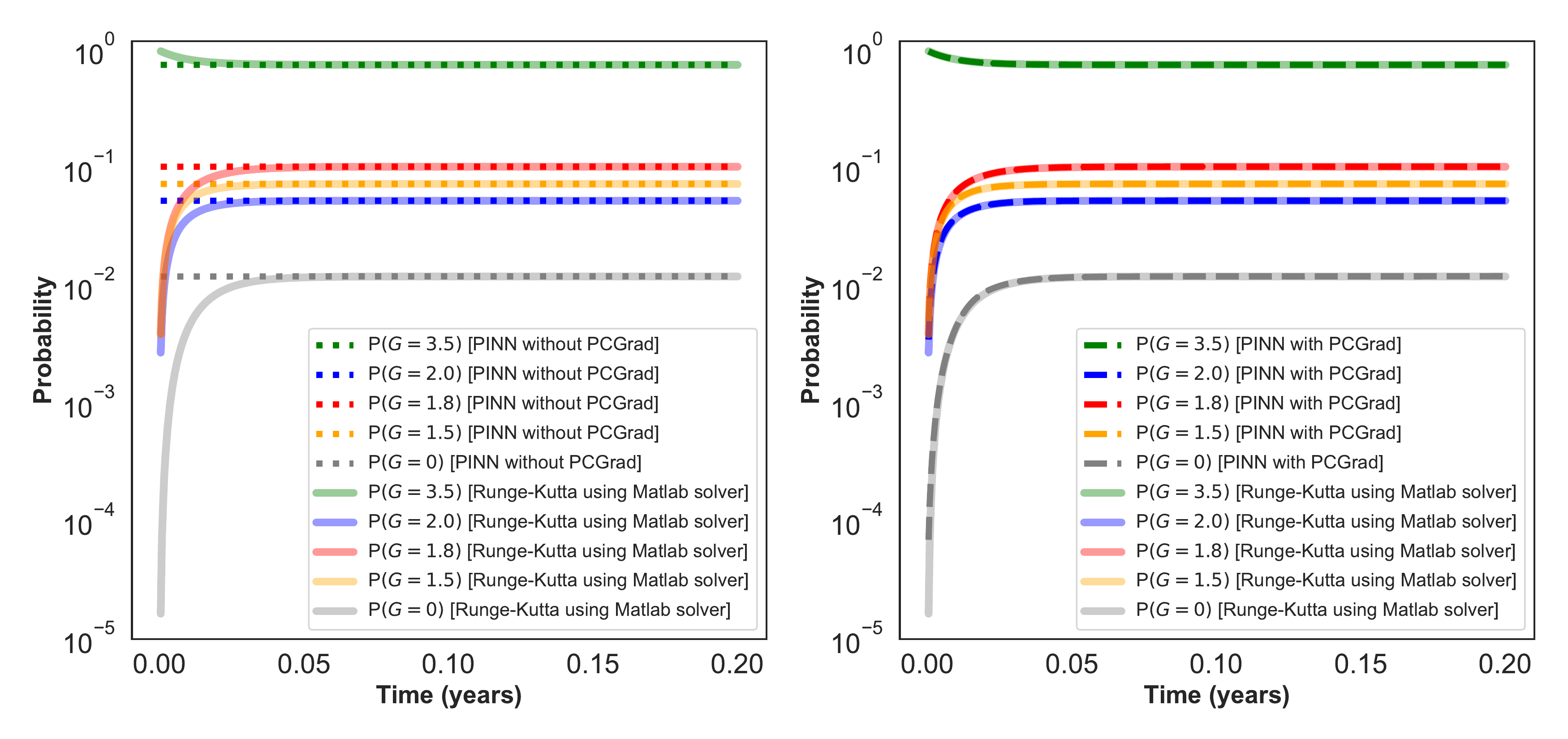}
    \caption{Instantaneous probabilities associated with different system performance rates in the flow transmission system}
    \label{fig:comparison_large}
\end{figure}

Fig. \ref{fig:comparison_large} visualizes the results of PINN when estimating the probabilities associated with each system performance rate in comparison with the solutions derived from the Matlab solver. Again, PINN with PCGrad achieves much better performance than the case of no PCGrad. In particular, when $t$ is less than 0.05, PINN without PCGrad fails to capture the changing trend of the probability associated with each system performance rate while PINN with PCGrad matches with the Matlab solver consistently when estimating the probability corresponding to each system performance rate.  

Next, we compute the RMSE of the differences in the solutions derived by PINN and the Matlab solver. Table~\ref{tab:performance_comparison_large_example} summarizes the RMSE between PINN and the Matlab solver in terms of the five system performance rates. Clearly, the RMSE of PINN with PCGrad is less than the RMSE of no PCGrad by nearly two orders of magnitude. In other words, PCGrad substantially improves the solution quality when performing reliability assessment in MSS. 
 
\begin{table}[!ht]
    \centering
    \caption{Performance comparisons of PINN with and without PCGrad against Matlab solver in the reliability assessment of the flow transmission system}
    \begin{tabular}{c|c}
    \toprule
     \textbf{Methods}    &  \textbf{Time Range}\\
    \midrule
    & $t \in \left[0, 0.2\right]$ \\
    \hline
     RMSE of PINN without PCGrad    & 0.013388 \\
     RMSE of PINN with PCGrad       & 0.000454 \\
    \bottomrule
    \end{tabular}
    \label{tab:performance_comparison_large_example}
\end{table}

\subsection{Summary}
As demonstrated in the previous three numerical examples, the benefits of deconflicting gradients using PCGrad in PINN are multi-fold. First of all, it substantially reduces the number of iterations and the amount of training data needed to tune the PINN, thus facilitating to achieve data and computation-efficient PINN. It also alleviates the oscillation of loss values during the training of PINN, and allows PINN to converge to a better solution with a much lower RMSE in relation to the solutions derived using the Matlab solver than that of PINN without PCGrad. Last but not least, the introduction of gradient projection frees us from tuning the weight parameter $\mathcal{W}$ as shown in Eq. (\ref{eq:mss_loss}) because all the tasks are treated independently in the PINN with PCGrad and the weight parameter $\mathcal{W}$ is not needed any more. 

\section{Conclusion}
Reliability assessment of multi-state systems is of significant concerns in a broad range of areas. In this paper, we exploit the power of physics-informed neural network and formulate a generic PINN-based framework for MSS reliability assessment. The developed framework tackles the problem of MSS reliability assessment from a machine learning perspective, and provides a viable paradigm for effective reliability modeling. The proposed methodology follows a two-step procedure. In the first step, MSS reliability assessment is reformulated as a machine learning problem in the framework of PINN, where loss functions are constructed to characterize the constraints associated with the initial condition and state transitions in MSS. Afterwards, to mitigate the high imbalance in the magnitude of gradients during the training of PINN, we leverage the projecting conflicting gradients (PCGrad) method to map the gradient of a task onto the norm plane of the other task that has a conflicting gradient. The embedding of PCGrad into the optimization algorithms significantly speeds up the convergence of the PINN to high-quality solutions. The proposed PINN-based framework demonstrates promising performance in evaluating the reliability of MSS in a variety of scenarios.

Future work can be carried out along the following directions. First of all, we investigate PINN's applications in the MSS, where state transitions are characterized by either homogeneous or non-homogeneous CTMC. It is worth exploring how to adopt PINN to analyze the reliability of semi-Markov MSS. Another direction worthy of investigation is to explore more effective ways to incorporate the equations governing state transitions in MSS into the neural network. In this paper, the ODE residuals are embedded into the neural network in a soft manner by appropriately penalizing the loss function. The drawback of this approach is that PINN might still violate the state transition equations in some scenarios. Thus, it is meaningful to investigate other alternative approaches so as to guarantee that PINN is strictly in compliance with the equations governing the underlying state transitions in MSS. Last but not least, we add the training points representing the ODE for state transitions at one shot, it is essential to develop more effective methods to add the ODE residual points in an adaptive manner, for example, adaptively adding training points in the locations with the largest expected reduction in the ODE loss in batch mode. 

\section*{Acknowledgement}
The work described in this paper is partially supported by a grant from the Research Committee of The Hong Kong Polytechnic University under project code 1-BE6V.

\bibliographystyle{elsarticle-num-names}
\bibliography{ref}

\end{document}